\documentclass{article} 
\usepackage{iclr2027_conference,times}

\usepackage{amsmath,amsfonts,bm}

\def\eqref#1{equation~\ref{#1}}

\def\1{\bm{1}}

\DeclareMathAlphabet{\mathsfit}{\encodingdefault}{\sfdefault}{m}{sl}
\SetMathAlphabet{\mathsfit}{bold}{\encodingdefault}{\sfdefault}{bx}{n}

\newif\ifpreprintversion
\preprintversiontrue

\usepackage{hyperref}
\usepackage{url}
\usepackage{xspace}
\usepackage{algorithm}
\usepackage{algpseudocode}
\usepackage{amssymb}
\usepackage{booktabs}
\usepackage[table]{xcolor}
\usepackage{tabularx}
\usepackage{array}
\usepackage{wrapfig}

\usepackage{listings}
\usepackage{tcolorbox}
\tcbuselibrary{skins}
\usepackage{placeins}

\ifpreprintversion
  \usepackage{tcolorbox}
  \tcbuselibrary{breakable,skins}

  \definecolor{abstractcardbackground}{HTML}{F3F8FA}
  \newcommand{\preprintnotice}{Preprint. Work in progress.}
  \newcommand{\projectcodeurl}{https://github.com/syr-cn/PluginRSI}
  \newcommand{\homepageurl}{https://syr-cn.github.io/PluginRSI/}

  \newtcolorbox{abstractcard}{
    enhanced,
    breakable,
    colback=abstractcardbackground,
    colframe=abstractcardbackground,
    boxrule=0pt,
    arc=3mm,
    outer arc=3mm,
    boxsep=0pt,
    left=6mm,
    right=6mm,
    top=4mm,
    bottom=4mm,
    before skip=0.5em,
    after skip=1em
  }

  \newenvironment{paperabstract}{%
    \begin{abstractcard}
    {\centering\large\scshape\color{black} Abstract\par}
    \vspace{0.5em}
  }{%
    \par\vspace{1.25ex}
    \noindent\color{black}
    \textbf{Date:} \today\par
    \noindent\textbf{Homepage:} \href{\homepageurl}{\texttt{\homepageurl}}\par
    \noindent\textbf{Code:} \href{\projectcodeurl}{\texttt{\projectcodeurl}}
    \end{abstractcard}
  }
  \usepackage{graphicx}
  \newcommand{\longcatlogofile}{figures/longcat-logo-full.pdf}
  \newcommand{\longcatlogo}[1]{%
    \AtPageUpperLeft{\hspace*{\dimexpr\paperwidth-1.5in\relax}%
      \llap{\raisebox{-13.2mm}{\includegraphics[height=#1]{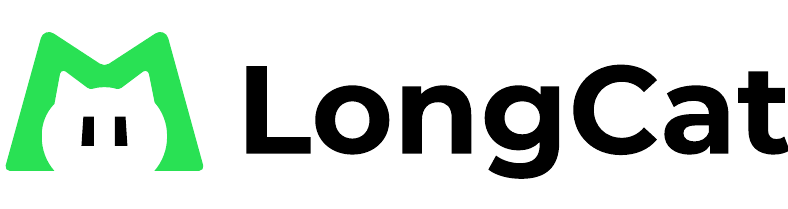}}}}}
  \AddToShipoutPictureFG{%
    \ifnum\value{page}=1\relax
      \longcatlogo{8mm}%
    \else
      \longcatlogo{5.8mm}%
    \fi
  }
\else
  \newenvironment{paperabstract}{\begin{abstract}}{Code is available at \url{https://anonymous.4open.science/r/PluginRSI-175A/}.\end{abstract}}
\fi

\newcommand{\plugharness}{PluginRSI\xspace}

\title{
\plugharness: Recursive Improvement of Agent Harnesses with Reusable Plugins
}

\author{
{\normalfont
    \begin{tabular}{c}
    \textbf{Yaorui Shi}\textsuperscript{1,2},
    \textbf{Yuchun Miao}\textsuperscript{2,3},
    \textbf{Yuxin Chen}\textsuperscript{2,4},
    \textbf{Jiayuan Zhang}\textsuperscript{2},
    \textbf{Yueqing Sun}\textsuperscript{2}, \\
    \textbf{Xierui Song}\textsuperscript{2,\ensuremath{\dagger}},
    \textbf{Xiang Wang}\textsuperscript{1},
    \textbf{An Zhang}\textsuperscript{1,\ensuremath{\dagger}} \\[0.5em]
    \textsuperscript{1}University of Science and Technology of China
    \quad \textsuperscript{2}Meituan \\
    \textsuperscript{3}Wuhan University
    \quad \textsuperscript{4}National University of Singapore
    \\[0.5em]
    \textsuperscript{\ensuremath{\dagger}}Corresponding authors:
    \texttt{guqi03@meituan.com}, \texttt{an\_zhang@ustc.edu.cn}
    \end{tabular}%
}%
}

\newcommand{\eg}{\emph{e.g., }}

\definecolor{observationbg}{HTML}{E8F6F6}

\newcommand{\observation}[2]{%
  \begin{tcolorbox}[
    colback=observationbg,
    colframe=observationbg,
    boxrule=0pt,
    arc=2mm,
    boxsep=0pt,
    left=8pt,
    right=8pt,
    top=6pt,
    bottom=6pt,
    before skip=8pt,
    after skip=8pt
  ]
    \textbf{Obs.~#1:}\enspace #2
  \end{tcolorbox}%
}

\usepackage{xcolor}
\usepackage{tcolorbox}
\tcbuselibrary{skins,breakable}

\definecolor{promptbackground}{HTML}{f4f8fa}
\definecolor{promptframe}{HTML}{1e5aa1}

\newtcolorbox{promptbox}[1]{
  enhanced,
  breakable,
  colback=promptbackground,
  colframe=promptframe,
  colbacktitle=promptframe,
  coltitle=white,
  fonttitle=\bfseries\small,
  fontupper=\small,
  title={#1},
  boxrule=0.6pt,
  arc=2mm,
  sharp corners=south,
  drop shadow,
  left=3mm,
  right=3mm,
  top=2mm,
  bottom=2mm,
  toptitle=1.5mm,
  bottomtitle=1.5mm,
  before skip=12pt,
  after skip=12pt,
  before upper={
    \setlength{\parindent}{0pt}
    \setlength{\parskip}{5pt}
  }
}

\ifpreprintversion
  \iclrfinalcopy
\fi
\begin{document}

\maketitle

\ifpreprintversion
  \lhead{\preprintnotice}
\fi

\begin{paperabstract}
The harness surrounding a language model is a central determinant of agent performance.
Recent methods optimize harnesses by searching over complete programs, where individual mechanisms are difficult to isolate and reuse.
We introduce \plugharness, which represents a harness as a composition of atomized plugins and organizes harness evolution around these plugins.
Individual plugins are improved independently and accumulated in a shared library, then recombined into new harnesses at each iteration.
\plugharness improves over existing harness optimization methods across software engineering, command-line interaction, and question-answering tasks.
The resulting harnesses retain their advantage when transferred to other solver models without further optimization.
The evolved plugin library accelerates subsequent optimization from the initial harness, which helps faster and higher convergence on unseen tasks.
These results show that accumulating reusable mechanisms provides an effective basis for continued harness improvement.
\end{paperabstract}

\begin{quote}
\itshape
\centering
Everything is a plugin. \\
\hfill ---- DeepSeek Harness
\end{quote}


\section{Introduction}
\label{sec:introduction}

Recursive self-improvement aims to enable large language model (LLM) agents to become more capable as they continue to operate~\citep{rsi_survey1, rsi_survey2, rsi_survey3}.
Such improvement can come from updating the underlying model parameters or refining the harness that organizes an agent's interaction mechanisms such as tools, roles and skills~\citep{FunctionsAsLearnableWeights, AgentHarnessSurvey, SkillClaw}.
Harness optimization pursues the latter, allowing agent behavior to be adapted without the training cost of weight updates~\citep{code_harness, nl_harness}.
Modern harnesses use plugin systems to implement new mechanisms as independent, reusable components, supporting continued improvement~\citep{deepseek-harness, ClaudeCode, Codex}.

Recent work explores recursive harness optimization by iteratively rewriting harness implementations based on execution feedback~\citep{adas,aflow,dgm,meta-harness, AgenticHarnessEngineering, ContinualHarness, SelfHarness}.
Despite its effectiveness, iterative harness rewriting faces two limitations.
First, it limits the exploration of new mechanisms.
Multiple mechanisms are coupled within the harness implementation, making it difficult to independently explore new mechanisms and incorporate their benefits into the harness.
Consequently, optimization could repeatedly converge to similar designs.
Second, valuable mechanisms are difficult to preserve and reuse.
Rewriting can overwrite effective local mechanisms, preventing improvements from accumulating into reusable experience across iterations.
Together, these limitations hinder continued improvement by restricting mechanism discovery and the accumulation of reusable mechanisms.

\begin{figure}[t]
    \centering
    \includegraphics[width=0.8\linewidth]{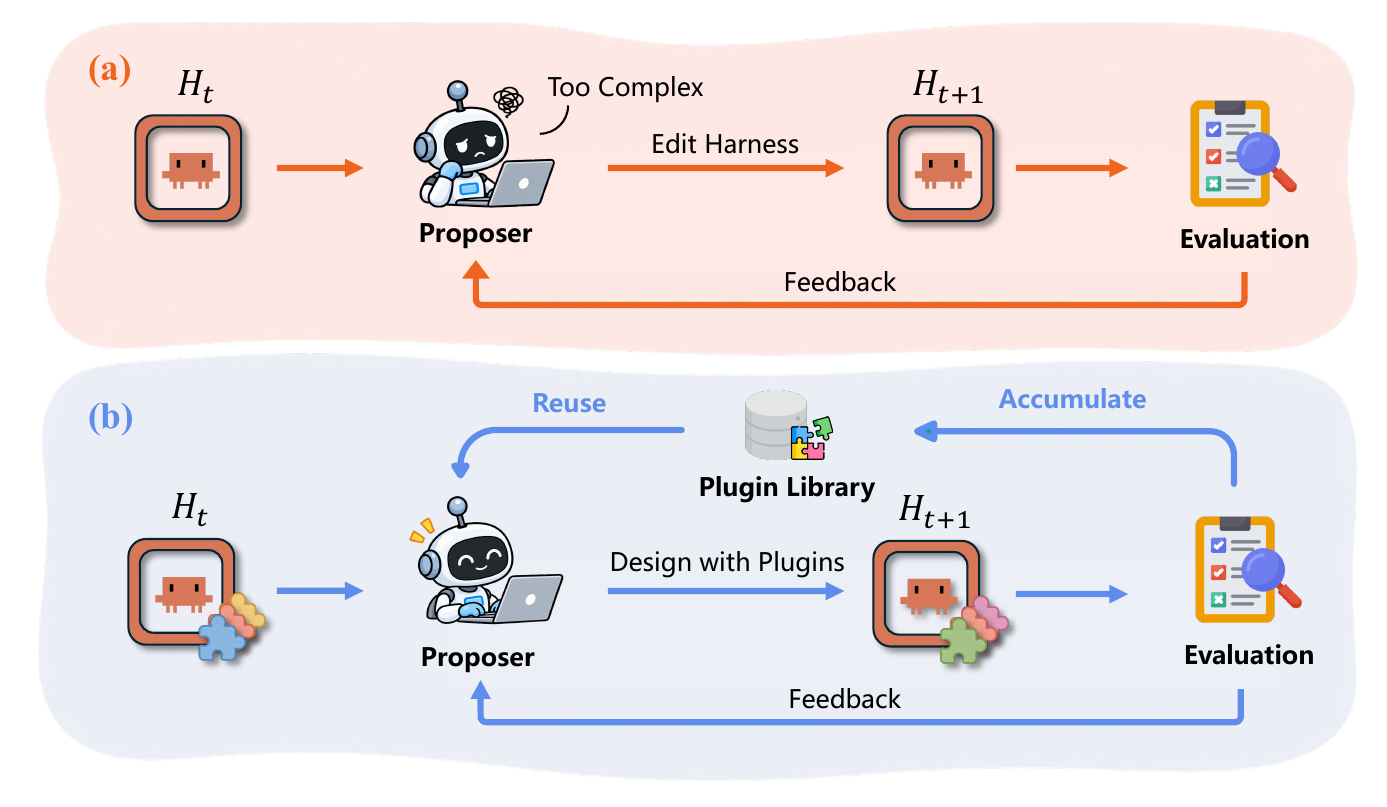}
    \vspace{-5pt}
    \caption{
        \textbf{Comparison of harness optimization paradigms.}
        \textbf{(a)} Prior approaches expose the complete harness $H_t$ as the object of modification.
        \textbf{(b)} \plugharness explores recursive optimization of plugin-parameterized harnesses, where each plugin encapsulates an independent mechanism that can be accumulated and reused.
    }
    \label{fig:paradigm}
\end{figure}

We argue that harness evolution should center on discovering new mechanisms and reusing them across candidates, rather than repeatedly rewriting harness implementations.
Inspired by this, we study the recursive optimization of \textbf{plugin-parameterized harnesses}.
As shown in Figure~\ref{fig:paradigm}, plugin-parameterized harnesses encapsulate individual mechanisms as plugins with standardized interfaces.
This formulation allows individual mechanisms to be developed and their contributions evaluated while keeping the rest of the harness unchanged.
A shared plugin library retains empirically beneficial implementations, allowing mechanisms discovered during optimization to accumulate and be reused to construct subsequent harness candidates.

Towards this end, we introduce \textbf{\plugharness}, an algorithm that enables \textbf{r}ecursive harne\textbf{s}s \textbf{i}mprovement by accumulating and reusing \textbf{plug}ins.
\plugharness features a shared plugin library that accumulates beneficial harness-design mechanisms as plugins.
Each harness consists of a set of plugins drawn from this library and a workflow that coordinates their execution.
One optimization iteration begins with a plugin mutation step, where execution feedback guides changes to the plugins used by a candidate harness, exploring new mechanisms while keeping the remaining harness fixed.
Plugin variants that improve harness performance are added to the library.
Harness recomposition then selects plugins from the updated library and revises the workflow to assemble new harness candidates, building on the mechanisms discovered in earlier iterations.

We evaluate \plugharness{} on software engineering, command-line interaction, and question-answering tasks.
Across these settings, \plugharness{} outperforms baseline methods.
On SWE-bench Verified, it improves held-out resolve rates over Meta-Harness by 6--10 percentage points across two optimization backbones.
The resulting harnesses retain their advantage across solver models without further optimization.
Optimization curves show that \plugharness{} finds strong harnesses with fewer solver rollouts.
Ablations demonstrate the contributions of both plugin mutation and harness recomposition, with strong performance retained without the initial plugin library.
The accumulated plugins support larger harness implementations with compact workflows and accelerate subsequent optimization.
Reusing the evolved library from the initial harness reaches a held-out resolve rate of 69\% after two evolution steps, compared with 65\% without reuse.

\begin{figure}[t]
    \centering
    \includegraphics[width=\linewidth]{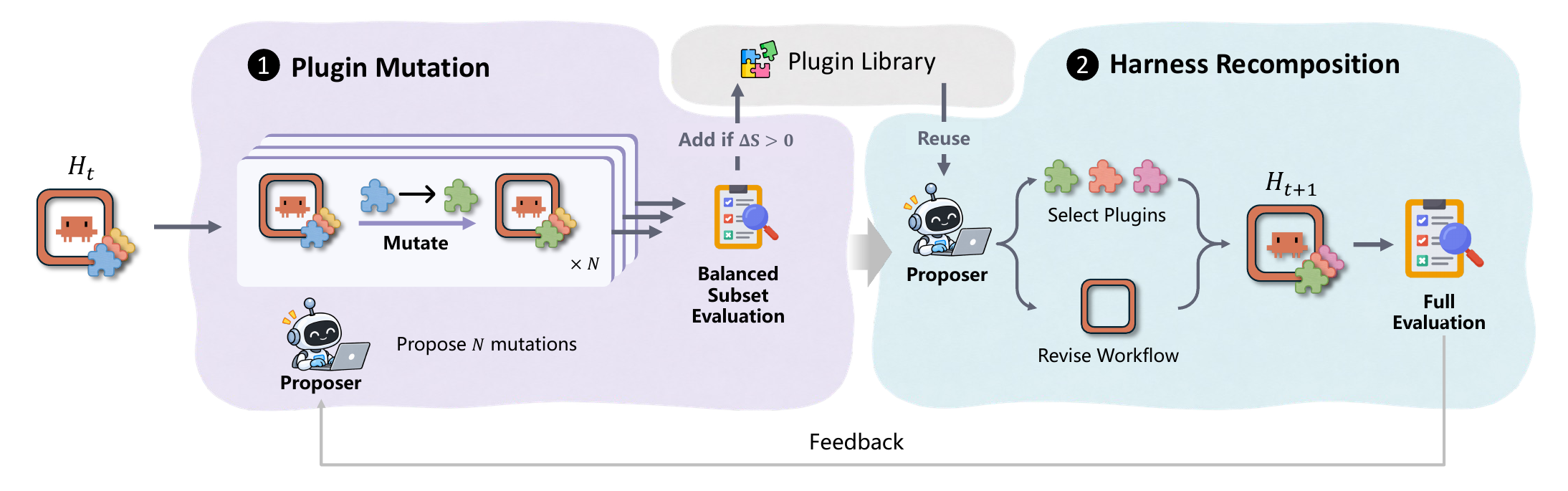}
    \vspace{-5pt}
    \caption{
        \textbf{Overview of \plugharness.}
        \textbf{(1) Plugin mutation} explores individual plugin changes in parallel while keeping the remaining harness fixed.
        \textbf{(2) Harness recomposition} selects plugins from the updated library and revises the workflow to construct a new harness candidate.
    }
    \vspace{-5pt}
    \label{fig:framework}
\end{figure}

\section{Problem Formulation}
\label{sec:preliminary}

\subsection{Optimization Objective}
\label{sec:preliminary-objective}

A harness specifies how an agent interacts with its environment by defining and coordinating tools, skills, and other mechanisms.
Together, a harness $H$ and a policy model $\Theta$ define an agent $\Phi(\cdot\mid H,\Theta)$.
Given a task $x$ sampled from a task distribution $\mathcal D$, the agent generates a trajectory $\tau\sim\Phi(\cdot\mid x,H,\Theta)$, whose performance is evaluated by a reward function $r(\tau,x)$.
We define the expected reward of a harness as
\begin{equation}
J(H)
=
\mathbb E_{\substack{x\sim\mathcal D,\tau\sim\Phi(\cdot\mid x,H,\Theta)}}
\left[r(\tau,x)\right].
\label{eq:optimize-objective}
\end{equation}
Harness optimization seeks a harness $H^\star\in\arg\max_H J(H)$ that maximizes Eq.~\ref{eq:optimize-objective} while keeping the policy model $\Theta$ fixed.

\subsection{Recursive Harness Optimization}
\label{sec:preliminary-loop}

Recursive harness optimization uses feedback from previously evaluated harnesses to guide the construction of new candidates.
A separate proposer agent $\Phi_{\mathrm{proposer}}$ examines a candidate's implementation and execution trajectories, reviews the agent's behavior, and proposes new candidates.
Both the policy model $\Theta$ and the proposer $\Phi_{\mathrm{proposer}}$ remain fixed throughout the search.

The search begins with an initial harness $H_0$.
We consider a search procedure in which, at iteration $t$, the proposer generates a new candidate based on the previous best harness and its feedback:
\begin{equation}
H_t\sim\Phi_{\mathrm{proposer}}\!\left(
\cdot\mid
\widehat H_t,\mathcal T_{\widehat H_t}
\right),
\label{eq:optimize-step}
\end{equation}
where $\widehat H_t\in\arg\max_{H\in\{H_i\}_{i=0}^{t-1}}J(H)$ is the best harness found so far, and $\mathcal T_{\widehat H_t}$ denotes the execution trajectories collected for $\widehat H_t$.
The agent defined by the proposed harness, $\Phi(\cdot\mid H_t,\Theta)$, is then evaluated on tasks sampled from $\mathcal D$, producing rewards and trajectories for subsequent iterations.
After $T$ iterations, the search returns $H^*\in\arg\max_{H\in\{H_i\}_{i=0}^{T}}J(H)$.

\section{\plugharness: Plugin-Oriented Harness Optimization}
\label{sec:method}

We introduce \plugharness, a recursive harness optimization algorithm that operates over plugin-parameterized harnesses.
The algorithm represents a harness through plugin implementations and the workflow that coordinates them, while maintaining a shared library that retains beneficial implementations across iterations (\S\ref{sec:method-representation}).

The optimization loop of \plugharness is described in \S\ref{sec:method-loop-init} and illustrated in Figure~\ref{fig:framework}.
At each iteration, \plugharness first explores new mechanisms through plugin mutation, where individual plugins are selected, mutated, and evaluated while the remaining harness is kept fixed (\S\ref{sec:method-loop-mutation}).
Plugin variants that yield empirical gains are accumulated in the shared plugin library.
The algorithm then performs harness recomposition, which selects plugins from the updated library and revises the workflow that coordinates them to explore new harness designs (\S\ref{sec:method-loop-recomposition}).

\subsection{Plugin-Parameterized Harnesses}
\label{sec:method-representation}

A plugin $p$ is an executable implementation of a harness mechanism exposed through a standardized interface.
This interface separates a mechanism's implementation from the workflow that invokes it.
In \plugharness, each plugin consists of two components: (1) a YAML specification defining its interface and associated configuration, including input and output schemas, prompts, rubrics, and metadata, and (2) an executable implementation written in \texttt{Python} code.
Further schema details are provided in Appendix~\ref{app:schema}.

Building on this abstraction, we represent a plugin-parameterized harness as $H=\mathcal H(W,\mathcal P)$, where $\mathcal P$ denotes the set of selected plugins and $W$ denotes the workflow.
The plugins in $\mathcal P$ implement the agent's roles, tools, skills, and memory mechanisms according to the schema.
The workflow $W$ determines when these plugins are invoked and how their outputs are used in subsequent interactions.
The function $\mathcal H$ is a deterministic program that compiles $W$ and $\mathcal{P}$ into an executable harness.

This representation supports two complementary optimization operations.
First, individual plugin implementations can be discovered or refined while keeping the workflow and the remaining plugins fixed.
Second, plugins can be selected from a plugin library and the workflow revised to coordinate their execution.
These operations correspond to the plugin mutation and harness recomposition stages of \plugharness, respectively, as discussed in \S\ref{sec:method-loop}.

\subsection{The Optimization Loop}
\label{sec:method-loop}

\subsubsection{Initialization and overview.}
\label{sec:method-loop-init}

The search starts from an initial harness $H=H_0=\mathcal H(W_0,\mathcal P_0)$ and a plugin library $\mathcal L$, with $\mathcal P_0\subseteq\mathcal L$.
The harness $H$ is iteratively evaluated on validation tasks $\mathcal D_{\mathrm{val}}$, and the scores and execution trajectories are used for analysis to propose a new harness $H'$.
Here $\mathcal D_{\mathrm{val}}$ and $\mathcal D_{\mathrm{test}}$ are disjoint validation and test datasets drawn from task distribution $\mathcal D$, where $\mathcal D_{\mathrm{test}}$ is held-out and not involved in the optimization process.

Each iteration consists of two stages.
The \textbf{plugin mutation} stage explores changes to individual plugins and adds beneficial implementations to $\mathcal L$.
\textbf{Harness recomposition} then selects plugins from the updated library and revises the workflow to construct a new candidate.
After $T$ iterations, the algorithm returns the best harness on $\mathcal D_\mathrm{val}$.
Algorithm~\ref{alg:plugharness} summarizes this process.

\subsubsection{Plugin mutation.}
\label{sec:method-loop-mutation}

Plugin mutation uses execution feedback to guide changes to individual mechanisms in the current harness.
At each iteration, we generate $N$ mutation candidates in parallel from $H$, with each candidate using a separately sampled minibatch $\mathcal M$ of $b$ tasks from $\mathcal D_{\mathrm{val}}$.
Based on the current harness's most recent evaluation results, each minibatch contains equal numbers of correctly and incorrectly solved tasks.
The separate sampling of minibatches provides varied execution feedback to encourage diverse mechanism exploration.

For each minibatch, the proposer agent examines $H$ and its execution feedback on that minibatch to select a plugin $p_i\in\mathcal P$ and produce a modified implementation $p_i'$.
The modified plugin replaces $p_i$ while the workflow and the remaining plugins stay fixed.
The resulting plugin set is compiled with the unchanged workflow to obtain $H_i=\mathcal H(W,(\mathcal P\setminus\{p_i\})\cup\{p_i'\})$.
We evaluate $H_i$ on its corresponding minibatch and compare with the score of $H$ on the same minibatch.
Plugin variants with empirical improvements are added to $\mathcal L$.

\subsubsection{Harness recomposition.}
\label{sec:method-loop-recomposition}

Harness recomposition uses the implementations accumulated in $\mathcal L$ to explore new harness designs.
This process allows both the plugin selection $\mathcal P$ and the workflow $W$ to change.

At this stage, the proposer examines previous harness execution feedback and the mutation results.
These comparisons guide the selection of a plugin set $\mathcal P'\subseteq\mathcal L$ that incorporates promising implementations from the updated library.
The proposer also revises the workflow from $W\rightarrow W'$ to improve the plugin coordination, including when they are invoked and how information flows.
The new harness candidate is then compiled as $H'=\mathcal H(W',\mathcal P')$ and evaluated on $\mathcal D_{\mathrm{val}}$.

\definecolor{alggray}{HTML}{777777}
\definecolor{algblue}{HTML}{2F6FDB}
\definecolor{alggreen}{HTML}{088943}

\newcommand{\AlgSection}[1]{%
    \State {\textcolor{alggreen}{\footnotesize\ttfamily // #1}}%
}
\algrenewcommand{\algorithmiccomment}[1]{%
    \hfill\textcolor{alggray}{\(\triangleright\) #1}%
}

\begin{algorithm}[t]
\caption{\plugharness: Plugin-Oriented Harness Optimization}
\label{alg:plugharness}
\begingroup
\linespread{1.08}\selectfont
\begin{algorithmic}[1]
\small
\Require Initial harness $H_0=\mathcal{H}(W_0,\mathcal{P}_0)$, proposer agent $\Phi_{\mathrm{proposer}}$, validation tasks $\mathcal{D}_{\mathrm{val}}$
\Require Minibatch size $b$, iterations $T$, mutation candidates per iteration $N$
\Ensure Optimized harness $H^*$

\State Initialize plugin library $\mathcal{L}$ and incumbent $(H,W,\mathcal{P}) \gets (H_0,W_0,\mathcal{P}_0)$
\State Evaluate initial $(S[H],\mathcal{T}_{H}) \gets \operatorname{Evaluate}(H,\mathcal{D}_{\mathrm{val}})$
\For{$t=1,\ldots,T$}
    \AlgSection{Step 1: Plugin Mutation}
    \For{$i=1,\ldots,N$ \textbf{in parallel}} \algorithmiccomment{parallel exploration on different subsets}
        \State $\mathcal{M} \gets$ balanced minibatch of $b$ tasks from $\mathcal{D}_{\mathrm{val}}$,
        \State Gather score $S_{\mathcal{M}}[H]$ and execution trace $\mathcal{T}_{H,\mathcal{M}}$ for $H$ on $\mathcal{M}$
        \State $\Phi_{\mathrm{proposer}}$ selects and mutates plugin $p_i\rightarrow p_i'$
        \State $H_i \gets \mathcal{H}\left(W,(\mathcal{P}\setminus\{p_i\})\cup\{p_i'\}\right)$ \algorithmiccomment{recompile with mutated plugin}
        \State Gather score $S_{\mathcal{M}}[H_i]$ and execution trace $\mathcal{T}_{H_i,\mathcal{M}}$ for $H_i$ on $\mathcal{M}$
        \State Add each $p_i'$ with $S_{\mathcal{M}}[H_i]>S_{\mathcal{M}}[H]$ to $\mathcal{L}$ \algorithmiccomment{accumulate beneficial plugins}
    \EndFor
    \AlgSection{Step 2: Harness Recomposition}
    \State $\Phi_{\mathrm{proposer}}$ revises workflow $W\rightarrow W'$ and selects plugins $\mathcal{P}'\subseteq\mathcal{L}$
    \State $H' \gets \mathcal{H}(W',\mathcal{P}')$ \algorithmiccomment{recompile with new workflow and plugins}
    \State Evaluate $H'$ on $\mathcal{D}_{\mathrm{val}}$ to obtain $(S[H'],\mathcal{T}_{H'})$
    \If{$S[H']>S[H]$}
        \State $(H,W,\mathcal{P}) \gets (H',W',\mathcal{P}')$
    \EndIf
\EndFor
\State \Return $H$
\end{algorithmic}
\endgroup
\end{algorithm}

\newcommand{\modelicon}[2]{\raisebox{-0.25\height}{\includegraphics[height=1.15em]{figures/model_icon/#1}}~#2}

\section{Experiments}
\label{sec:experiments}

\subsection{Experimental Setup}
\label{sec:experimental-setup}

\paragraph{Benchmarks and protocol.}
We evaluate on software engineering tasks from SWE-bench Verified~\citep{swebench-verified}; command line interaction tasks from Terminal-Bench-2.1~\citep{terminalbench}; and Question-Answering (QA) tasks from SuperGPQA~\citep{supergpqa}, SciCite~\citep{scicite}, FinanceQA~\citep{financeqa}, FrontierScience~\citep{frontierscience} and OlympiadBench~\citep{olympiadbench}.
We split SWE-bench Verified into 400 held-in and 100 held-out tasks, and Terminal-Bench-2.1 into 59 held-in and 30 held-out tasks, as detailed in Appendix~\ref{app:dataset-split}.
For QA, we sample 50 tasks from each of eight domains with random seed 50.
We optimize on 150 tasks from Olympiad-Math, SciCite, and SuperGPQA-Economics, and evaluate on the other five domains.
For each optimization method, we select the harness with the highest held-in score and fix it for subsequent evaluation.
We report resolve rates (\%) for SWE-bench and Terminal-Bench, and accuracies for QA.

\paragraph{Backbone models.}
We separately optimize SWE-Bench Verified harnesses with GPT-5.6 Terra~\citep{gpt56} and Kimi-K3~\citep{kimi-k3} as proposers.
We use Kimi-K3 as the solver for Terminal-Bench and QA.
We then conduct evaluation on three different base models GPT-5.6 Terra, Kimi-K3 and GLM-5.2~\citep{glm-5} to measure cross-model utility.
By default, reasoning effort is set to \texttt{xhigh} for the proposer and \texttt{none} for the solver.
We also evaluate each selected harness with the other solver backbones, keeping its workflow and plugins fixed.
Full model configurations are provided in Appendix~\ref{app:implmentation_details}.

\paragraph{Baselines.}
We compare with two groups of methods.
\textbf{(1) Expert-curated harnesses} include ReAct~\citep{ReAct}, which follows a reasoning--acting loop in zero-shot and few-shot settings; and ACE~\citep{ace}, which accumulates experience in a context playbook.
\textbf{(2) Automatically optimized harnesses} include GEPA~\citep{gepa}, which performs reflective evolution at the prompt level; and Darwin-Godel Machine (DGM)~\citep{dgm} and Meta-Harness~\citep{meta-harness}, which search over complete harness implementations.
All methods share the same task splits, API endpoints, and environment sandboxes.

\begin{table}[t]
  \centering
  \caption{
    Comparison on SWE-bench Verified~\citep{swebench-verified}.
    Harnesses are optimized and selected on 400 held-in tasks, then fixed for evaluation across solver backbones.
    Values are resolve rates (\%).
    $\dagger$ marks the score used for harness selection.
    Bold marks the best result in each column within each optimization setting.
  }
  \label{tab:main-results}
  \vspace{5pt}
  \small
  \renewcommand{\arraystretch}{1.02}
  \resizebox{\linewidth}{!}{
  \begin{tabular}{lccc ccc}
    \toprule
    & \multicolumn{3}{c}{Held-In (400)}
    & \multicolumn{3}{c}{Held-Out (100)} \\
    \cmidrule(lr){2-4}\cmidrule(lr){5-7}
    Method
      & \modelicon{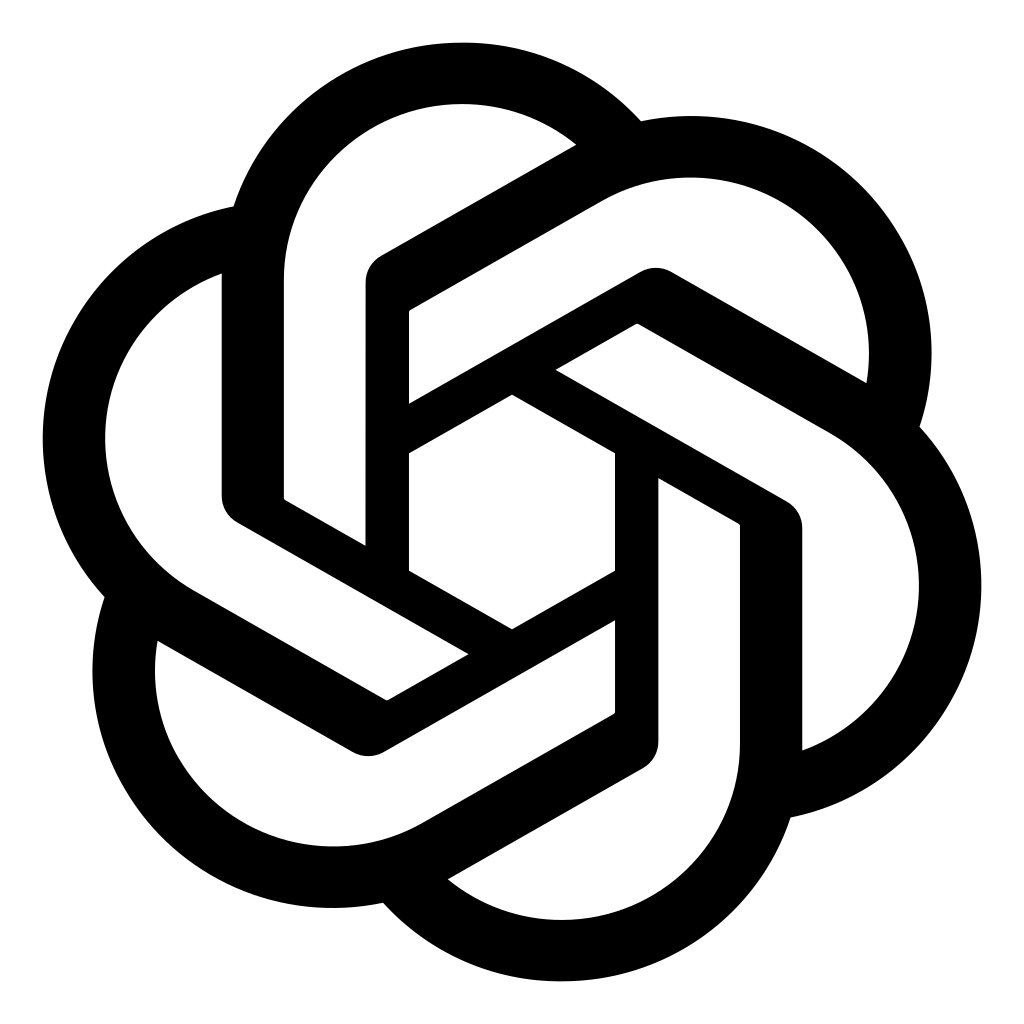}{GPT-5.6 Terra} & \modelicon{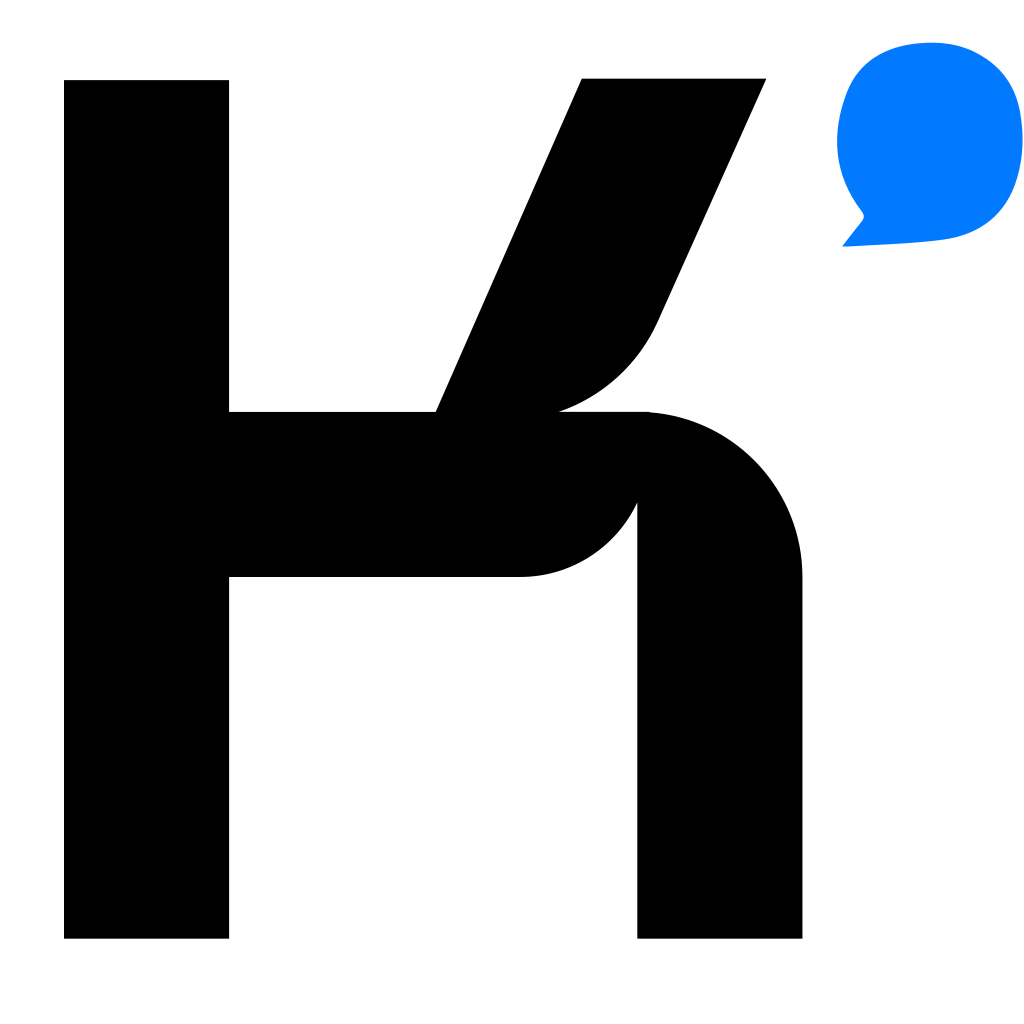}{Kimi-K3} & \modelicon{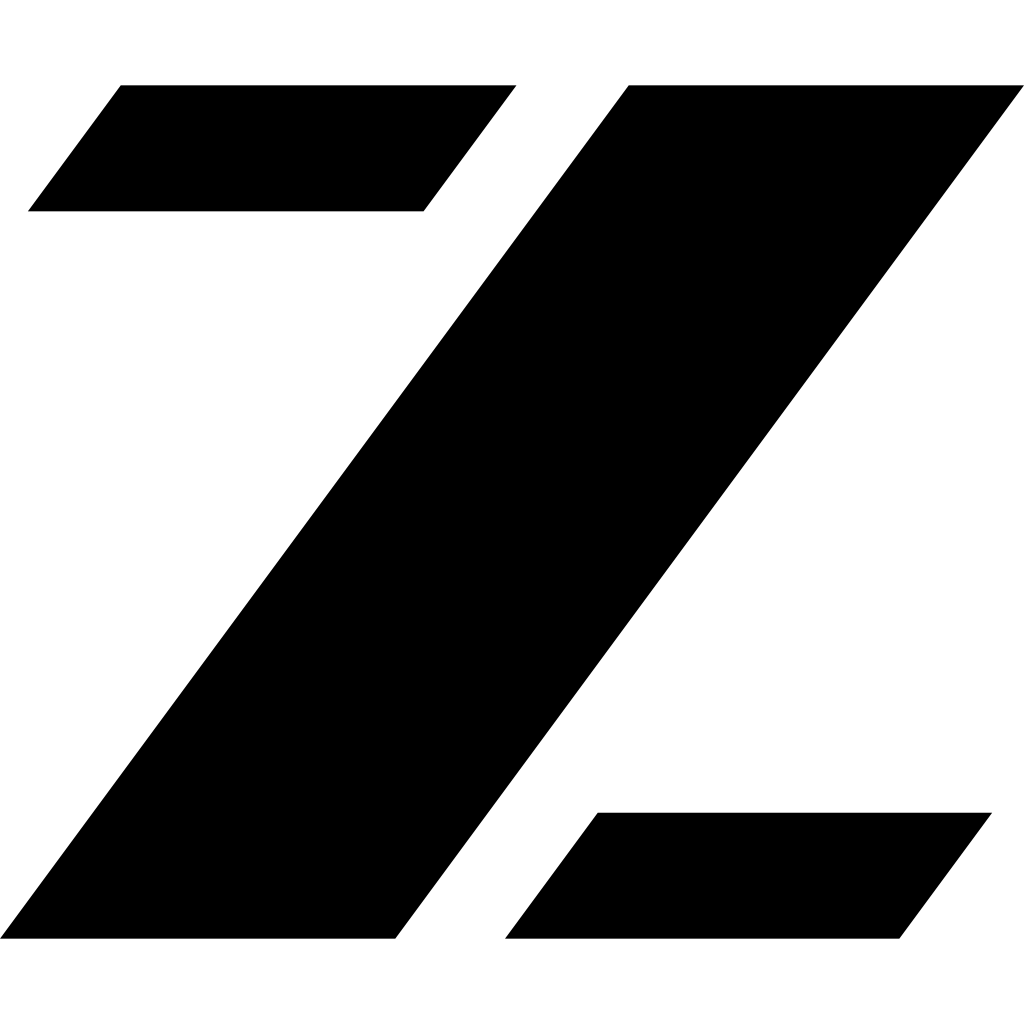}{GLM-5.2}
      & \modelicon{gpt.png}{GPT-5.6 Terra} & \modelicon{kimi.png}{Kimi-K3} & \modelicon{zai.png}{GLM-5.2} \\
    \midrule

    \rowcolor{gray!12}
    \multicolumn{7}{c}{\textit{Expert-Curated Harnesses}} \\

    ReAct (Zero-Shot)
      & 38.25 & 41.00 & 37.75
      & 39.00 & 41.00 & 41.00 \\

    ReAct (Few-Shot)
      & 52.50 & 44.25 & 38.25
      & 55.00 & 47.00 & 41.00 \\

    ACE
      & 46.75 & 41.00 & 30.25
      & 43.00 & 37.00 & 31.00 \\

    \midrule
    \rowcolor{gray!12}
    \multicolumn{7}{c}{\textit{Automatically Optimized Harnesses w/ GPT-5.6 Terra}} \\

    GEPA
      & 50.50$^\dagger$ & 42.50 & 39.25
      & 51.00 & 44.00 & 48.00 \\

    DGM
      & 49.00$^\dagger$ & 42.00 & 37.25
      & 50.00 & 51.00 & 43.00 \\

    Meta-Harness
      & 54.50$^\dagger$ & 54.00 & 45.25
      & 55.00 & 53.00 & 44.00 \\

    \textbf{\plugharness}
      & \textbf{58.50}$^\dagger$ & \textbf{60.50} & \textbf{48.00}
      & \textbf{65.00} & \textbf{60.00} & \textbf{52.00} \\

    \midrule
    \rowcolor{gray!12}
    \multicolumn{7}{c}{\textit{Automatically Optimized Harnesses w/ Kimi-K3}} \\

    GEPA
      & 55.00 & 61.00$^\dagger$ & 47.00
      & 61.00 & 62.00 & 52.00 \\

    DGM
      & 57.75 & 58.00$^\dagger$ & 50.00
      & 58.00 & 55.00 & 51.00 \\

    Meta-Harness
      & \textbf{58.25} & 64.25$^\dagger$ & \textbf{59.75}
      & 60.00 & 63.00 & 59.00 \\

    \textbf{\plugharness}
      & 58.00 & \textbf{66.75}$^\dagger$ & 59.50
      & \textbf{64.00} & \textbf{69.00} & \textbf{64.00} \\

    \bottomrule
  \end{tabular}
  }
\end{table}

\subsection{Main Results}
\label{sec:main-results}

\paragraph{Results on SWE-bench Verified.}
Table~\ref{tab:main-results} shows that \plugharness{} improves performance with both optimization backbones.
We observe higher held-in results for both GPT- and Kimi-optimized harnesses.
With GPT-5.6 Terra and Kimi-K3, it achieves held-out resolve rates of 65.00\% and 69.00\%, exceeding the best baseline by 10.00 and 6.00 percentage points, respectively.
These gains are larger than the corresponding held-in gains of 4.00 and 2.50 points.
The discovered improvements therefore remain effective beyond the tasks used for harness selection.

\begin{wraptable}{r}{0.5\textwidth}
  \centering
  \vspace{-15pt}
  \caption{
    Results on Terminal-Bench-2.1 with Kimi-K3.
    Values are resolve rates (\%).
    Bold marks the best result in each column.
  }
  \label{tab:terminal-results}
  \vspace{5pt}
  \small
  \setlength{\tabcolsep}{4pt}
  \renewcommand{\arraystretch}{1.05}
  \resizebox{0.9\linewidth}{!}{
    \begin{tabular}{@{}lcc@{}}
      \toprule
      Method & Held-In (59) & Held-Out (30) \\
      \midrule
      ReAct (Zero-Shot) & 47.5 & 50.0 \\
      ACE              & 54.2 & 56.7 \\
      GEPA             & 57.6 & 63.3 \\
      DGM              & 57.6 & 60.0 \\
      Meta-Harness     & 64.4 & 66.7 \\
      \textbf{\plugharness}
        & \textbf{69.5} & \textbf{73.3} \\
      \bottomrule
    \end{tabular}
  }
  \vspace{0pt}
\end{wraptable}

\paragraph{Transfer across backbone models.}
The optimized harnesses retain their advantage when transferred to other solver models without further optimization.
The harness optimized with GPT-5.6 Terra achieves held-out resolve rates of 60.00\% with Kimi-K3 and 52.00\% with GLM-5.2, exceeding the strongest baselines by 7.00 and 4.00 points.
The harness optimized with Kimi-K3 achieves 64.00\% with both GPT-5.6 Terra and GLM-5.2, improving over the strongest baselines by 3.00 and 5.00 points.
For some held-in cross-model evaluations we also observe a performance degradation against baseline methods, \eg 58.00\% against previous best 58.25\%, where the performance gaps remain modest.
We regard these as comparable results.

\begin{table}[t]
  \centering
  \caption{
    Results on QA benchmarks.
    We optimize harnesses on 150 tasks from three held-in domains.
    Both optimization and evaluation are on Kimi-K3.
    Values are accuracies rounded to 3 decimals.
    Bold indicates the best results.
  }
  \label{tab:qa-results}
  \vspace{5pt}
  \small
  \renewcommand{\arraystretch}{1.05}
  \resizebox{\linewidth}{!}{
  \begin{tabular}{lcccc cccccc c}
    \toprule
    & \multicolumn{4}{c}{Held-In (150)}
    & \multicolumn{6}{c}{Held-Out (250)}
    & \\
    \cmidrule(lr){2-5}\cmidrule(lr){6-11}
    Method
      & Math & SciCite & Eco & Avg.
      & Fin & FS & Physics & Law & Med & Avg.
      & Overall \\
    \midrule

    ReAct
      & 0.480 & 0.540 & 0.420 & 0.480
      & 0.280 & 0.620 & 0.140 & 0.320 & 0.300 & 0.332
      & 0.388 \\

    GEPA
      & 0.680 & 0.640 & 0.580 & 0.633
      & 0.380 & 0.660 & 0.200 & 0.620 & 0.500 & 0.472
      & 0.533 \\

    Meta-Harness
      & 0.780 & 0.680 & \textbf{0.800} & 0.753
      & 0.480 & \textbf{0.740} & \textbf{0.280} & 0.800 & 0.540 & 0.568
      & 0.638 \\

    \textbf{\plugharness}
      & \textbf{0.880} & \textbf{0.780} & \textbf{0.800} & \textbf{0.820}
      & \textbf{0.560} & 0.720 & \textbf{0.280} & \textbf{0.820} & \textbf{0.560} & \textbf{0.588}
      & \textbf{0.675} \\

    \bottomrule
  \end{tabular}
  }
\end{table}

\paragraph{Results on Terminal-Bench.}
Table~\ref{tab:terminal-results} shows that the benefits extend to command-line interaction tasks.
With Kimi-K3, \plugharness{} achieves 69.5\% on held-in tasks and 73.3\% on held-out tasks, compared with 64.4\% and 66.7\% for Meta-Harness.
The improvements are larger on held-out tasks.

\paragraph{Results on QA benchmarks.}
Table~\ref{tab:qa-results} shows the results.
Math and Physics are from OlympiadBench~\citep{olympiadbench}; Eco(nomics), Law, and Med(icine) are from SuperGPQA~\citep{supergpqa}.
FS stands for FrontierScience~\citep{frontierscience} and Fin stands for FinanceQA~\citep{financeqa}.
\plugharness{} improves held-in accuracy from 0.753 to 0.820 and overall accuracy from 0.638 to 0.675 over baselines.
Held-out accuracy increases more modestly from 0.568 to 0.588.
The smaller held-out gains are expected given the limited overlap among QA domains, as different domains require different specialized strategies.
For example, strategies discovered for mathematical reasoning may not directly transfer to law questions.

\observation{1}{
\plugharness{} improves harness performance across task settings. The advantages are retained across solver models, and are larger on unseen similar tasks.
}

\begin{figure}[t]
    \centering
    \includegraphics[width=\linewidth]{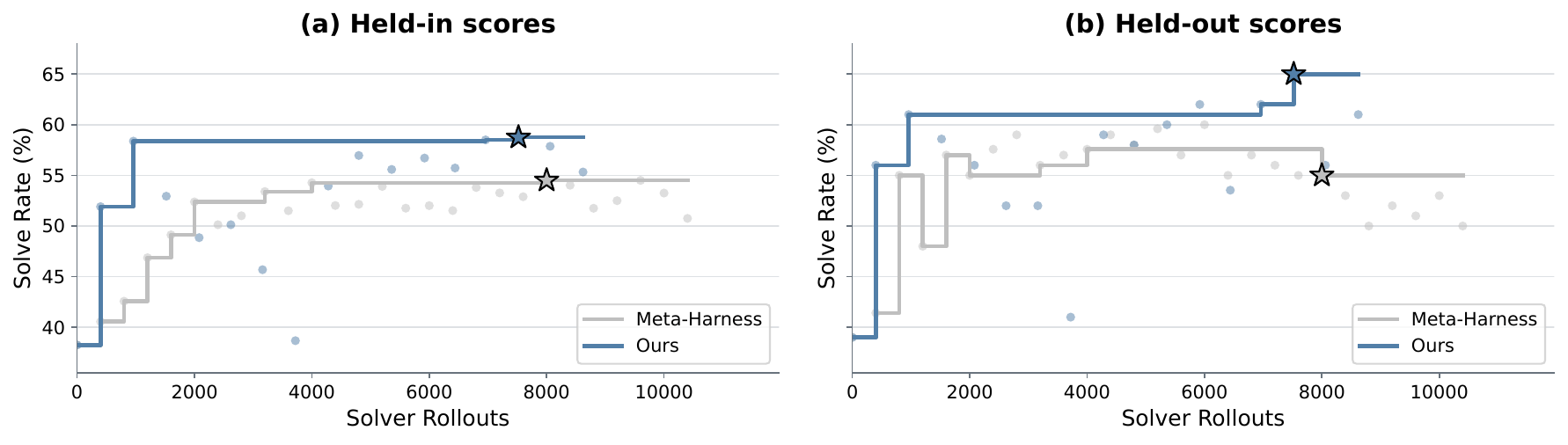}
    \vspace{-10pt}
    \caption{
        \textbf{Optimization dynamics.}
        Held-in and held-out resolve rates of the best held-in harness against cumulative solver rollouts.
        \plugharness yields higher performing harness with fewer solver rollouts, and held-in performances aligns better with held-out scores.
    }
    \label{fig:analysis-dynamic}
\end{figure}

\subsection{Optimization Dynamics}
\label{sec:analysis-dynamics}

We examine how efficiently \plugharness{} improves harness performance.
Figure~\ref{fig:analysis-dynamic} compares \plugharness{} with Meta-Harness using GPT-5.6 Terra.
We plot the best held-in scores and the held-out scores of the corresponding harnesses against cumulative solver rollouts.
The x-axis is the number of solver rollouts for fair comparison, since the plugin mutation stage introduces extra burden.
Appendix~\ref{app:token-cost} provides a token cost analysis.

\plugharness{} rapidly reaches a held-in resolve rate that Meta-Harness does not attain within the evaluated budget.
Harnesses with better held-in performance also improve on held-out tasks, with early gains followed by further improvements in later iterations.
Meta-Harness improves more gradually on held-in tasks, while its held-out performance fluctuates and remains lower at the end of optimization.

\observation{2}{
\plugharness{} finds higher-performing harnesses with fewer solver rollouts, with held-in improvements accompanied by held-out gains.
}

\subsection{Ablation Study}
\label{sec:ablation}

We examine the contributions of the plugin library and the two optimization stages.
Table~\ref{tab:ablation} reports results under Kimi-K3 optimization and transfer to GPT-5.6 Terra and GLM-5.2.
For the \textit{w/o initial plugins} variant, we retain only the four basic plugins from the 75-plugin library and keep the initial workflow unchanged.
For the \textit{w/o harness recomposition} variant, we adopt the mutation-stage harness with the highest score on its $b$-task minibatch.

\paragraph{Plugin library and initial plugins.}
Adding an initialized or empty plugin library to Meta-Harness gives Kimi-K3 held-out resolve rates of 64.00\% and 62.00\%, close to its original 63.00\%.
This suggests that providing a plugin library alone yields limited gains.
Meanwhile, removing the initial plugins from \plugharness{} yields similar Kimi-K3 held-in scores (66.75\% versus 67.00\%), and reduces held-out scores by only 1.00--2.00 points across the three solvers.
Strong performance is retained without these initial plugins.

\paragraph{Plugin mutation and harness recomposition.}
Removing plugin mutation or harness recomposition reduces the Kimi-K3 held-out resolve rate from 69.00\% to 57.00\% and 59.00\%, respectively.
Both ablations also reduce held-out scores with GPT-5.6 Terra and GLM-5.2.
These results support combining local improvements to plugins with global changes to their coordination.

\observation{3}{
Both plugin mutation and harness recomposition improve held-out performance.
A plugin library alone does not bring obvious gains.
}

\begin{table}[t]
  \centering
  \caption{
    Ablation study on SWE-bench Verified.
    Harnesses are optimized and selected with Kimi-K3 on 400 held-in tasks, then fixed for evaluation across solver backbones.
    Values are resolve rates (\%) and $\dagger$ marks the score used for harness selection.
    Bold marks the best result in each column.
  }
  \label{tab:ablation}
  \vspace{5pt}
  \small
  \renewcommand{\arraystretch}{1.02}
  \resizebox{\linewidth}{!}{
  \begin{tabular}{lccc ccc}
    \toprule
    & \multicolumn{3}{c}{Held-In (400)}
    & \multicolumn{3}{c}{Held-Out (100)} \\
    \cmidrule(lr){2-4}\cmidrule(lr){5-7}
    Method
      & \modelicon{gpt.png}{GPT-5.6 Terra} & \modelicon{kimi.png}{Kimi-K3}$^\dagger$ & \modelicon{zai.png}{GLM-5.2}
      & \modelicon{gpt.png}{GPT-5.6 Terra} & \modelicon{kimi.png}{Kimi-K3} & \modelicon{zai.png}{GLM-5.2} \\
    \midrule

    Meta-Harness
      & 58.25 & 64.25 & \textbf{59.75}
      & 60.00 & 63.00 & 59.00 \\

    \quad + Plugin Library
      & 58.25 & 65.50 & 55.25
      & 58.00 & 64.00 & 56.00 \\

    \quad + Plugin Library w/o Initial Plugins
      & 61.00 & 64.75 & 56.25
      & 62.00 & 62.00 & 58.00 \\

    \midrule

    \textbf{\plugharness}
      & 58.00 & 66.75 & 59.50
      & \textbf{64.00} & \textbf{69.00} & \textbf{64.00} \\

    \quad w/o Initial Plugins
      & \textbf{62.25} & \textbf{67.00} & 57.75
      & 62.00 & 67.00 & 63.00 \\

    \quad w/o Plugin Mutation
      & 52.25 & 58.50 & 46.25
      & 49.00 & 57.00 & 50.00 \\

    \quad w/o Harness Recomposition
      & 57.25 & 62.50 & 59.00
      & 57.00 & 59.00 & 60.00 \\

    \bottomrule
  \end{tabular}
  }
\end{table}

\subsection{Plugin Accumulation and Reuse}
\label{sec:analysis-scale}

We examine how plugins accumulate during optimization and whether the evolved library accelerates subsequent harness optimization.

\begin{figure}[t]
    \centering

    \begin{minipage}[t]{0.48\linewidth}
        \centering
        \includegraphics[
            width=\linewidth,
        ]{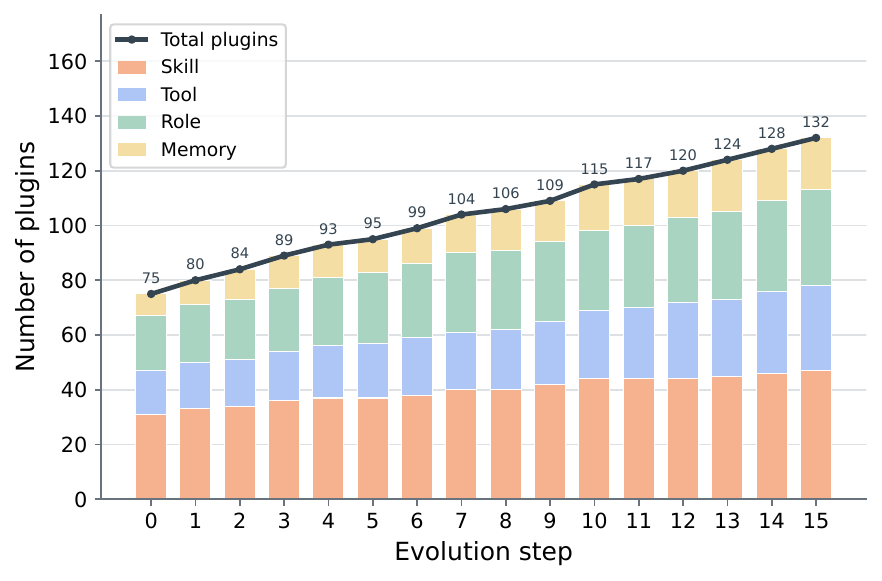}
        \par\vspace{5pt}
        {\small (a) Number of Plugins}
    \end{minipage}
    \hfill
    \begin{minipage}[t]{0.48\linewidth}
        \centering
        \includegraphics[
            width=\linewidth,
        ]{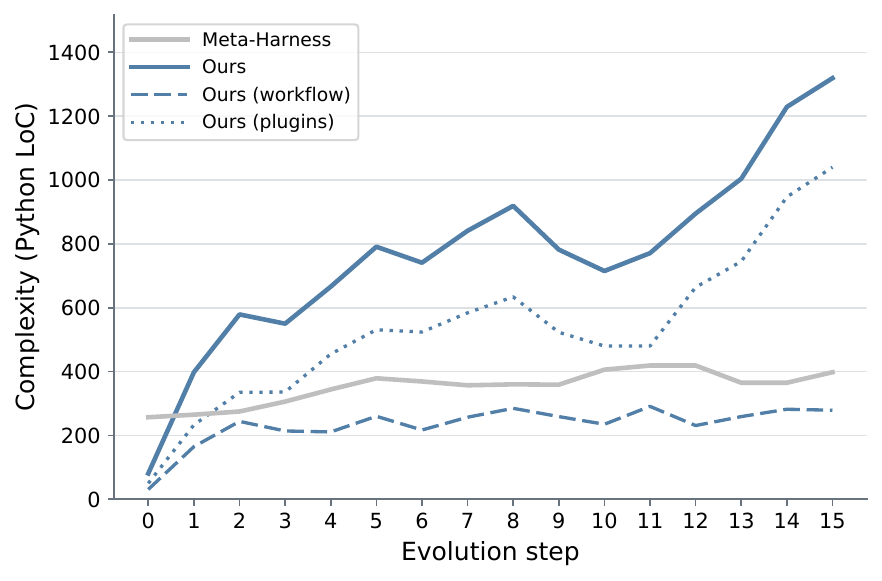}
        \par\vspace{5pt}
        {\small (b) Harness implementation size}
    \end{minipage}

    \caption{
        \textbf{Plugin accumulation statistics.}
        (a) Number of plugins in the plugin library.
        (b) Candidate implementation size during optimization. \plugharness explores more complex harness structure by reusing discovered plugins, while the workflow implementation remains relatively brief.
    }
    \label{fig:analysis-accumulation}
\end{figure}

\paragraph{Plugin accumulation.}
Figure~\ref{fig:analysis-accumulation}(a) shows that the library grows from 75 to 132 plugins over 15 evolution steps, with additions across all four categories.
Figure~\ref{fig:analysis-accumulation}(b) tracks candidate size in \texttt{Python} lines of code (LoC).
By the final step, \plugharness candidates reach approximately 1,300 LoC, compared with 400 LoC for Meta-Harness.
Most of this growth comes from plugin implementations, while the workflow remains below 300 LoC.
These results show that accumulated plugins support larger harness implementations while keeping the workflow compact.

\begin{wrapfigure}{h}{0.48\linewidth}
    \centering
    \vspace{-25pt}
    \includegraphics[width=\linewidth]{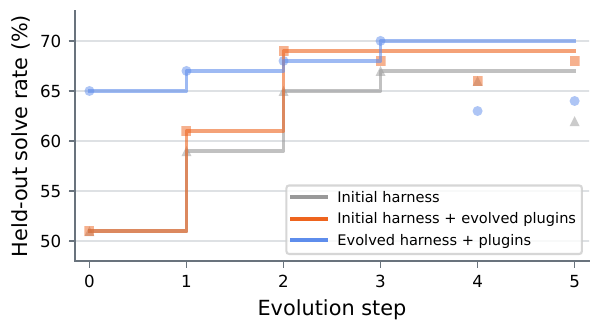}
    \vspace{-20pt}
    \caption{\textbf{Plugin reuse effectiveness.} On held-out tasks, optimizing from initial harness w/ the evolved plugin library accelerates convergence.}
    \vspace{-10pt}
    \label{fig:analysis-reuse}
\end{wrapfigure}

\paragraph{Reusing evolved plugins.}
Figure~\ref{fig:analysis-reuse} compares optimization from the initial harness, the initial harness with the evolved library, and the evolved harness with its plugins.
Reusing the library from the initial harness reaches a held-out resolve rate of 69\% after two evolution steps, compared with 65\% without reuse.
This approaches the 70\% reached after three steps of continued optimization from the evolved harness.
The results show that the evolved library preserves useful improvements that accelerate optimization from a different starting harness.

\observation{4}{
Accumulated plugins support larger harness implementations with compact workflows.
Reusing the evolved library accelerates subsequent optimization.
}

\section{Related Work}
\label{sec:related-work}

\textbf{Self-Improving Agent.}
Existing work enables agents to improve from feedback through model parameter updates and the refinement of external artifacts.
Policy learning approaches let agents themselves automatically generate tasks and interaction trajectories, and be optimized via reinforcement learning~\citep{absolute0, r0, webrl,agentevolver,agent0} or supervised fine-tuning~\citep{seal}.
For artifact optimization, prompt optimizers update instructions using execution feedback~\citep{human-level-prompt-engineers,textgrad,gepa}, while context and memory methods organize information and experiences for subsequent tasks~\citep{ace,reasoningbank,rememr1,reflexion,expel,MemoryR1, memocr}.
Other approaches acquire and retain reusable behaviors as agent skills~\citep{Voyager,SkillRL,SkillClaw}.
\plugharness studies the improvement of executable mechanisms and their coordinating workflows via harness optimization.

\textbf{Recursive Harness Optimization.}
Recursive harness optimization improves agent performance by searching over harness designs.
Existing methods explore this space by iteratively rewriting harness implementations.
One line of work targets specific aspects of the harness, including agentic workflows~\citep{adas,aflow}, prompts and role assignments~\citep{gepa,optimize_anything}, and executable components~\citep{skillopt,skilladaptor,mcp-zero, skill1}.
Other studies directly optimize algorithms~\citep{math_discovery,alphaevolve} and executable implementations of agents~\citep{dgm,Hyperagents,meta-harness,SelfHarness,ContinualHarness,AdaptiveAutoHarness}.
However, mechanisms remain entangled within the harness during iterative rewriting, making their individual contributions difficult to evaluate and preventing their reuse.
\plugharness uses a plugin system to decouple individual mechanisms, enabling their evaluation with the rest of the harness fixed and their reuse across candidates.

\textbf{Modular Harness Design.}
Expert-curated agent systems adopt modular harness designs to organize human-designed mechanisms into composable components.
For example, DeepSeek Harness exposes runtime capabilities as swappable plugins~\citep{deepseek-harness}.
This modular organization allows human-designed mechanisms to be incorporated into harness engineering~\citep{AgentSquare,lego-rl,legomem}.
Recent work~\citep{HarnessX, harnessbank} combines typed processors to automatically edit harness components and configurations.
\plugharness explores recursive harness optimization of plugin-parameterized harnesses, focusing on how plugins can be independently discovered, accumulated and reused across iterations.


\section{Conclusion}
We introduce \plugharness{}, a plugin-oriented algorithm for recursive harness optimization.
It improves individual mechanisms through plugin mutation and recomposes harnesses from an accumulated plugin library.
Experiments show performance gains across software engineering, command-line interaction, and QA tasks.
On SWE-bench Verified, \plugharness{} finds stronger harnesses with fewer solver rollouts and retains its held-out advantage across solver models.
Ablations support the contributions of both optimization stages.
Reusing the evolved library accelerates subsequent optimization from the initial harness.
These results show that harness evolution can accumulate reusable mechanisms that support further improvement.


\bibliography{iclr2027_conference}

@inproceedings{ace,
    title={Agentic Context Engineering: Evolving Contexts for Self-Improving Language Models},
    author={Qizheng Zhang and Changran Hu and Shubhangi Upasani and Boyuan Ma and Fenglu Hong and Vamsidhar Kamanuru and Jay Rainton and Chen Wu and Mengmeng Ji and Hanchen Li and Urmish Thakker and James Zou and Kunle Olukotun},
    booktitle={The Fourteenth International Conference on Learning Representations},
    year={2026},
}

@misc{deepseek-harness,
  title={DeepSeek Harness: Everything Is a Plugin},
  author={{DeepSeek-AI}},
  howpublished={\url{https://deepseek.com/harness/en/}},
  year={2026},
  note={Accessed: 2026-09-22}
}

@article{skillopt,
  title={Skillopt: Executive strategy for self-evolving agent skills},
  author={Yang, Yifan and Gong, Ziyang and Huang, Weiquan and Yang, Qihao and Zhou, Ziwei and Huang, Zisu and Li, Yan and Gao, Xuemei and Dai, Qi and Liu, Bei and others},
  journal={arXiv preprint arXiv:2605.23904},
  year={2026}
}

@article{a-evolve,
  title={Position: Agentic evolution is the path to evolving llms},
  author={Lin, Minhua and Lu, Hanqing and Shi, Zhan and He, Bing and Mao, Rui and Zhang, Zhiwei and Wu, Zongyu and Tang, Xianfeng and Liu, Hui and Dai, Zhenwei and others},
  journal={arXiv preprint arXiv:2602.00359},
  year={2026}
}

@article{skilladaptor,
  title={Skilladaptor: Self-adapting skills for llm agents from trajectories},
  author={Yu, Zhuoyun and Xie, Xin and Yao, Wuguannan and Wang, Chenxi and Liang, Lei and Qi, Xiang and Deng, Shumin},
  journal={arXiv preprint arXiv:2606.01311},
  year={2026}
}

@inproceedings{rememr1,
  title={Look back to reason forward: Revisitable memory for long-context llm agents},
  author={Shi, Yaorui and Chen, Yuxin and Wang, Siyuan and Li, Sihang and Cai, Hengxing and Gu, Qi and Wang, Xiang and Zhang, An},
  booktitle={International Conference on Learning Representations},
  volume={2026},
  pages={78055--78082},
  year={2026}
}

@inproceedings{memocr,
    title={Mem{OCR}: Layout-Aware Visual Memory for Efficient Long-Horizon Reasoning},
    author={Yaorui Shi and Shugui Liu and Yu Yang and Wenyu Mao and Yuxin Chen and Qi GU and Hui Su and Xunliang Cai and Xiang Wang and An Zhang},
    booktitle={Forty-third International Conference on Machine Learning},
    year={2026},
}

@article{skill1,
  title={Skill1: Unified evolution of skill-augmented agents via reinforcement learning},
  author={Shi, Yaorui and Chen, Yuxin and Lu, Zhengxi and Miao, Yuchun and Liu, Shugui and Gu, Qi and Cai, Xunliang and Wang, Xiang and Zhang, An},
  journal={arXiv preprint arXiv:2605.06130},
  year={2026}
}

@inproceedings{aflow,
    title={{AF}low: Automating Agentic Workflow Generation},
    author={Jiayi Zhang and Jinyu Xiang and Zhaoyang Yu and Fengwei Teng and Xiong-Hui Chen and Jiaqi Chen and Mingchen Zhuge and Xin Cheng and Sirui Hong and Jinlin Wang and Bingnan Zheng and Bang Liu and Yuyu Luo and Chenglin Wu},
    booktitle={The Thirteenth International Conference on Learning Representations},
    year={2025},
}

@misc{swebench-verified,
  title={Introducing {SWE}-bench Verified},
  author={Chowdhury, Neil and Aung, James and Shern, Chan Jun and Jaffe, Oliver and Sherburn, Dane and Starace, Giulio and Mays, Evan and Dias, Rachel and Aljubeh, Marwan and Glaese, Mia and Jimenez, Carlos E. and Yang, John and Ho, Leyton and Patwardhan, Tejal and Liu, Kevin and Madry, Aleksander},
  year={2024},
  url={https://openai.com/index/introducing-swe-bench-verified/},
  note={Accessed: 2026-09-22}
}

@misc{terminalbench,
      title={Terminal-Bench: Benchmarking Agents on Hard, Realistic Tasks in Command Line Interfaces},
      author={Mike A. Merrill and Alexander G. Shaw and Nicholas Carlini and Boxuan Li and Harsh Raj and Ivan Bercovich and Lin Shi and Jeong Yeon Shin and Thomas Walshe and E. Kelly Buchanan and Junhong Shen and Guanghao Ye and Haowei Lin and Jason Poulos and Maoyu Wang and Marianna Nezhurina and Jenia Jitsev and Di Lu and Orfeas Menis Mastromichalakis and Zhiwei Xu and Zizhao Chen and Yue Liu and Robert Zhang and Leon Liangyu Chen and Anurag Kashyap and Jan-Lucas Uslu and Jeffrey Li and Jianbo Wu and Minghao Yan and Song Bian and Vedang Sharma and Ke Sun and Steven Dillmann and Akshay Anand and Andrew Lanpouthakoun and Bardia Koopah and Changran Hu and Etash Guha and Gabriel H. S. Dreiman and Jiacheng Zhu and Karl Krauth and Li Zhong and Niklas Muennighoff and Robert Amanfu and Shangyin Tan and Shreyas Pimpalgaonkar and Tushar Aggarwal and Xiangning Lin and Xin Lan and Xuandong Zhao and Yiqing Liang and Yuanli Wang and Zilong Wang and Changzhi Zhou and David Heineman and Hange Liu and Harsh Trivedi and John Yang and Junhong Lin and Manish Shetty and Michael Yang and Nabil Omi and Negin Raoof and Shanda Li and Terry Yue Zhuo and Wuwei Lin and Yiwei Dai and Yuxin Wang and Wenhao Chai and Shang Zhou and Dariush Wahdany and Ziyu She and Jiaming Hu and Zhikang Dong and Yuxuan Zhu and Sasha Cui and Ahson Saiyed and Arinbjörn Kolbeinsson and Jesse Hu and Christopher Michael Rytting and Ryan Marten and Yixin Wang and Alex Dimakis and Andy Konwinski and Ludwig Schmidt},
      year={2026},
      eprint={2601.11868},
      archivePrefix={arXiv},
      primaryClass={cs.SE},
}

@article{supergpqa,
  title={Supergpqa: Scaling llm evaluation across 285 graduate disciplines},
  author={Du, Xeron and Yao, Yifan and Ma, Kaijing and Wang, Bingli and Zheng, Tianyu and Liu, Minghao and Liang, Yiming and Jin, Xiaolong and Wei, Zhenlin and Zheng, Chujie and others},
  journal={Advances in Neural Information Processing Systems},
  volume={38},
  year={2026}
}

@inproceedings{scicite,
    title = "Structural Scaffolds for Citation Intent Classification in Scientific Publications",
    author = "Cohan, Arman  and
      Ammar, Waleed  and
      van Zuylen, Madeleine  and
      Cady, Field",
    editor = "Burstein, Jill  and
      Doran, Christy  and
      Solorio, Thamar",
    booktitle = "Proceedings of the 2019 Conference of the North {A}merican Chapter of the Association for Computational Linguistics: Human Language Technologies, Volume 1 (Long and Short Papers)",
    month = jun,
    year = "2019",
    address = "Minneapolis, Minnesota",
    publisher = "Association for Computational Linguistics",
    doi = "10.18653/v1/N19-1361",
    pages = "3586--3596",
}

@article{financeqa,
  title={Financeqa: A benchmark for evaluating financial analysis capabilities of large language models},
  author={Mateega, Spencer and Georgescu, Carlos and Tang, Danny},
  journal={arXiv preprint arXiv:2501.18062},
  year={2025}
}

@article{frontierscience,
  title={FrontierScience: Evaluating AI's Ability to Perform Expert-Level Scientific Tasks},
  author={Wang, Miles and Lin, Robi and Hu, Kat and Jiao, Joy and Chowdhury, Neil and Chang, Ethan and Patwardhan, Tejal},
  journal={arXiv preprint arXiv:2601.21165},
  year={2026}
}

@inproceedings{olympiadbench,
  title={Olympiadbench: A challenging benchmark for promoting agi with olympiad-level bilingual multimodal scientific problems},
  author={He, Chaoqun and Luo, Renjie and Bai, Yuzhuo and Hu, Shengding and Thai, Zhen and Shen, Junhao and Hu, Jinyi and Han, Xu and Huang, Yujie and Zhang, Yuxiang and others},
  booktitle={Proceedings of the 62nd Annual Meeting of the Association for Computational Linguistics (Volume 1: Long Papers)},
  pages={3828--3850},
  year={2024}
}

@article{lego-rl,
  title={LEGO-RL: Harness-Native Reinforcement Learning for Coding Agents},
  author={Du, Yiming and Jiang, Yuxin and Yuan, Tao and Dai, Jianbo and Wang, Shaowei and Chen, Jierun and Tao, Chaofan and Yu, Xianzhi and Shang, Lifeng and Wong, Kam-Fai and others},
  journal={arXiv preprint arXiv:2608.17393},
  year={2026}
}

@article{legomem,
  title={Legomem: Modular procedural memory for multi-agent llm systems for workflow automation},
  author={Han, Dongge and Couturier, Camille and Diaz, Daniel Madrigal and Zhang, Xuchao and R{\"u}hle, Victor and Rajmohan, Saravan},
  journal={arXiv preprint arXiv:2510.04851},
  year={2025}
}

@inproceedings{gepa,
  title={Gepa: Reflective prompt evolution can outperform reinforcement learning},
  author={Agrawal, Lakshya A and Tan, Shangyin and Soylu, Dilara and Ziems, Noah and Khare, Rishi and Opsahl-Ong, Krista and Singhvi, Arnav and Shandilya, Herumb and Ryan, Michael J and Jiang, Meng and others},
  booktitle={International Conference on Learning Representations},
  volume={2026},
  pages={8479--8565},
  year={2026}
}

@article{human-level-prompt-engineers,
  title={Large language models are human-level prompt engineers},
  author={Zhou, Yongchao and Muresanu, Andrei Ioan and Han, Ziwen and Paster, Keiran and Pitis, Silviu and Chan, Harris and Ba, Jimmy},
  journal={arXiv preprint arXiv:2211.01910},
  year={2022}
}

@inproceedings{optimize_anything,
  title={optimize\_anything: A universal api for optimizing any text parameter},
  author={Agrawal, Lakshya A and Lee, Donghyun and Tan, Shangyin and Ma, Wenjie and Elmaaroufi, Karim and Seshia, Sanjit A. and Sen, Koushik and Klein, Dan and Stoica, Ion and Gonzalez, Joseph E. and Khattab, Omar and Dimakis, Alexandros G. and Zaharia, Matei},
  booktitle={Proceedings of the ACM Conference on AI and Agentic Systems},
  pages={1300--1304},
  year={2026}
}

@inproceedings{meta-harness,
  title     = {{Meta-Harness}: End-to-End Optimization of Model Harnesses},
  author    = {Lee, Yoonho and Nair, Roshen and Zhang, Qizheng and Lee, Kangwook and Khattab, Omar and Finn, Chelsea},
  booktitle = {Conference on Language Modeling},
  year      = {2026},
}

@inproceedings{SWEBench,
  title  = {Swe-bench: Can language models resolve real-world github issues?},
  author = {Jimenez, Carlos E and Yang, John and Wettig, Alexander and Yao, Shunyu and Pei, Kexin and Press, Ofir and Narasimhan, Karthik},
  year   = {2024},
  booktitle = {In International Conference on Learning Representations, volume 2024, pages 54107–54157},
}

@inproceedings{SWEAgent,
  title  = {SWE-agent: agent-computer interfaces enable automated software engineering},
  author = {Yang, J. and Jimenez, C. E. and Wettig, A. and Lieret, K. and Yao, S. and Narasimhan, K. R. and Press, O.},
  year   = {2024},
  booktitle = {The Thirty-Eighth Annual Conference on Neural Information Processing Systems},
  url    = {https://openreview.net/forum?id=mXpq6ut8J3\&referrer=\%5Bthe\%20profile\%20of\%20Shunyu\%20Yao\%5D(\%2Fprofile\%3Fid\%3D~Shunyu\_Yao1)},
}

@article{alphaevolve,
  title={Alphaevolve: A coding agent for scientific and algorithmic discovery},
  author={Novikov, Alexander and V{\~u}, Ng{\^a}n and Eisenberger, Marvin and Dupont, Emilien and Huang, Po-Sen and Wagner, Adam Zsolt and Shirobokov, Sergey and Kozlovskii, Borislav and Ruiz, Francisco JR and Mehrabian, Abbas and others},
  journal={arXiv preprint arXiv:2506.13131},
  year={2025}
}

@inproceedings{dgm,
  title     = {Darwin {G}{\"o}del Machine: Open-Ended Evolution of Self-Improving Agents},
  author    = {Zhang, Jenny and Hu, Shengran and Lu, Cong and Lange, Robert Tjarko and Clune, Jeff},
  booktitle = {The Fourteenth International Conference on Learning Representations},
  year      = {2026},
}

@article{mcp-zero,
  title={Mcp-zero: Active tool discovery for autonomous llm agents},
  author={Fei, Xiang and Zheng, Xiawu and Feng, Hao},
  journal={arXiv preprint arXiv:2506.01056},
  year={2025}
}

@inproceedings{SkillRL,
    title={Skill{RL}: Evolving Agents via Recursive Skill-Augmented Reinforcement Learning},
    author={Peng Xia and Jianwen Chen and Hanyang Wang and Jiaqi Liu and Kaide Zeng and Yu Wang and Siwei Han and Yiyang Zhou and Xujiang Zhao and Haifeng Chen and Zeyu Zheng and Cihang Xie and Huaxiu Yao},
    booktitle={ICLR 2026 Workshop on Lifelong Agents: Learning, Aligning, Evolving},
    year={2026},
}

@misc{ClaudeCode,
  title  = {Claude code},
  author = {{Anthropic}},
  year   = {2025},
  howpublished = {https://github.com/anthropics/claude-code, 2025},
  url    = {https://github.com/anthropics/claude-code},
}

@article{glm-5,
  title={Glm-5: from vibe coding to agentic engineering},
  author={Zeng, Aohan and Lv, Xin and Hou, Zhenyu and Du, Zhengxiao and Zheng, Qinkai and Chen, Bin and Yin, Da and Ge, Chendi and Huang, Chenghua and Xie, Chengxing and others},
  journal={arXiv preprint arXiv:2602.15763},
  year={2026}
}

@article{kimi-k3,
  title={Kimi k3: Open frontier intelligence},
  author={Team, Kimi and Bai, Tongtong and Bai, Yifan and Bao, Yiping and Cai, Jianfeng and Cai, Xinyuan and Cao, Peizhou and Cao, Yuxuan and Chai, Ziwei and Charles, Y and others},
  journal={arXiv preprint arXiv:2607.24653},
  year={2026}
}

@misc{gpt56,
  title        = {{GPT-5.6: Frontier intelligence that scales with your ambition}},
  author       = {{OpenAI}},
  year         = {2026},
  url          = {https://openai.com/index/gpt-5-6/}
}

@article{code_harness,
  title={Code as agent harness},
  author={Ning, Xuying and Tieu, Katherine and Fu, Dongqi and Wei, Tianxin and Li, Zihao and Bei, Yuanchen and Zou, Jiaru and Ai, Mengting and Liu, Zhining and Li, Ting-Wei and others},
  journal={arXiv preprint arXiv:2605.18747},
  year={2026}
}

@article{nl_harness,
  title={Natural-language agent harnesses},
  author={Pan, Linyue and Zou, Lexiao and Guo, Shuo and Ni, Jingchen and Zheng, Hai-Tao},
  journal={arXiv preprint arXiv:2603.25723},
  year={2026}
}

@article{AgenticHarnessEngineering,
  title   = {Agentic Harness Engineering: Observability-Driven Automatic Evolution of Coding-Agent Harnesses},
  author  = {Lin, Jiahang and Liu, Shichun and Pan, Chengjun and Lin, Lizhi and Dou, Shihan and Xi, Zhiheng and Huang, Xuanjing and Yan, Hang and Han, Zhenhua and Gui, Tao and Jiang, Yu-Gang},
  journal = {arXiv preprint arXiv:2604.25850},
  year    = {2026},
}

@article{harnessbank,
  title={Self-Evolving Agent Harnesses via Gated Semantic Quality-Diversity},
  author={Luo, Xiaotian and Wang, Fengxingyu and Hu, Chuanrui and Xue, Dizhan and Deng, Yafeng},
  journal={arXiv preprint arXiv:2607.13683},
  year={2026}
}

@article{textgrad,
  title={Textgrad: Automatic" differentiation" via text},
  author={Yuksekgonul, Mert and Bianchi, Federico and Boen, Joseph and Liu, Sheng and Huang, Zhi and Guestrin, Carlos and Zou, James},
  journal={arXiv preprint arXiv:2406.07496},
  year={2024}
}

@article{absolute0,
  title={Absolute zero: Reinforced self-play reasoning with zero data},
  author={Zhao, Andrew and Wu, Yiran and Wu, Tong and Xu, Quentin and Yue, Yang and Lin, Matthieu and Wang, Shenzhi and Wu, Qingyun and Zheng, Zilong and Huang, Gao},
  journal={Advances in Neural Information Processing Systems},
  volume={38},
  pages={105816--105879},
  year={2026}
}

@inproceedings{agent0,
    title={Agent0: Unleashing Self-Evolving Agents from Zero Data via Tool-Integrated Reasoning},
    author={Peng Xia and Kaide Zeng and Jiaqi Liu and Can Qin and Fang Wu and Yiyang Zhou and Caiming Xiong and Huaxiu Yao},
    booktitle={Third Conference on Language Modeling},
    year={2026},
}

@inproceedings{r0,
  title={R-zero: Self-evolving reasoning llm from zero data},
  author={Huang, Chengsong and Yu, Wenhao and Wang, Xiaoyang and Zhang, Hongming and Li, Zongxia and Li, Ruosen and Huang, Jiaxin and Mi, Haitao and Yu, Dong},
  booktitle={International Conference on Learning Representations},
  volume={2026},
  pages={130770--130790},
  year={2026}
}

@inproceedings{webrl,
  title={Webrl: Training llm web agents via self-evolving online curriculum reinforcement learning},
  author={Qi, Zehan and Liu, Xiao and Iong, Iat Long and Lai, Hanyu and Sun, Xueqiao and Sun, Jiadai and Yang, Xinyue and Yang, Yu and Yao, Shuntian and Xu, Wei and others},
  booktitle={International Conference on Learning Representations},
  volume={2025},
  pages={79791--79821},
  year={2025}
}

@article{agentevolver,
  title={Agentevolver: Towards efficient self-evolving agent system},
  author={Zhai, Yunpeng and Tao, Shuchang and Chen, Cheng and Zou, Anni and Chen, Ziqian and Fu, Qingxu and Mai, Shinji and Yu, Li and Deng, Jiaji and Cao, Zouying and others},
  journal={arXiv preprint arXiv:2511.10395},
  year={2025}
}

@article{seal,
  title={Self-adapting language models},
  author={Zweiger, Adam and Pari, Jyo and Guo, Han and Kim, Yoon and Agrawal, Pulkit},
  journal={Advances in Neural Information Processing Systems},
  volume={38},
  pages={74084--74115},
  year={2026}
}

@article{swesmith,
  title={Swe-smith: Scaling data for software engineering agents},
  author={Yang, John and Lieret, Kilian and Jimenez, Carlos and Wettig, Alexander and Khandpur, Kabir and Zhang, Yanzhe and Hui, Binyuan and Press, Ofir and Schmidt, Ludwig and Yang, Diyi},
  journal={Advances in Neural Information Processing Systems},
  volume={38},
  year={2026}
}

@article{agentless,
  title={Agentless: Demystifying llm-based software engineering agents},
  author={Xia, Chunqiu Steven and Deng, Yinlin and Dunn, Soren and Zhang, Lingming},
  journal={arXiv preprint arXiv:2407.01489},
  year={2024}
}

@article{AgentHarnessSurvey,
  title  = {Agent Harness for Large Language Model Agents: A Survey},
  author = {Meng, Q. and Wang, Y. and Chen, L. and Li, Y. and Wu, W. and Jiang, W. and Wang, Q. and Lu, C. and Gao, Y. and Wu, Y. and Hu, Y.},
  year   = {2026},
  journal = {Preprints},
  doi    = {10.20944/preprints202604.0428.v3},
}

@inproceedings{adas,
  title     = {Automated Design of Agentic Systems},
  author    = {Hu, Shengran and Lu, Cong and Clune, Jeff},
  booktitle = {The Thirteenth International Conference on Learning Representations},
  year      = {2025},
}

@misc{Codex,
  title  = {Codex},
  author = {{OpenAI}},
  year   = {2026},
  howpublished = {Note: Product page},
  url    = {https://openai.com/codex/},
}

@article{ContinualHarness,
  title={Continual harness: Online adaptation for self-improving foundation agents},
  author={Karten, Seth and Zhang, Joel and Upaa Jr, Tersoo and Feng, Ruirong and Li, Wenzhe and Shi, Chengshuai and Jin, Chi and Vodrahalli, Kiran},
  journal={arXiv preprint arXiv:2605.09998},
  year={2026}
}

@inproceedings{expel,
    author       = {Andrew Zhao and Daniel Huang and Quentin Xu and Matthieu Lin and Yong-Jin Liu and Gao Huang},
    title        = {ExpeL: LLM Agents Are Experiential Learners},
    booktitle    = {Thirty-Eighth {AAAI} Conference on Artificial Intelligence},
    year         = {2024},
    pages        = {19632--19642},
    publisher    = {{AAAI} Press},
    doi          = {10.1609/aaai.v38i17.29936}
}

@article{HarnessX,
  title   = {{HarnessX}: A Composable, Adaptive, and Evolvable Agent Harness Foundry},
  author  = {Chen, Tingyang and Lu, Shuo and Zhao, Kang and Meng, Weicheng and Teng, Hanlin and Li, Tianhao and Li, Chao and Liu, Xule and Liang, Jian and Zhang, Zhizhong and Xie, Yuan and Qu, Heng and Shao, Kun and Luan, Jian},
  journal = {arXiv preprint arXiv:2606.14249},
  year    = {2026},
}

@article{Hyperagents,
  title   = {{Hyperagents}},
  author  = {Zhang, Jenny and Zhao, Bingchen and Yang, Wannan and Foerster, Jakob and Clune, Jeff and Jiang, Minqi and Devlin, Sam and Shavrina, Tatiana},
  journal = {arXiv preprint arXiv:2603.19461},
  year    = {2026},
}

@article{math_discovery,
  title={Mathematical discoveries from program search with large language models},
  author={Romera-Paredes, Bernardino and Barekatain, Mohammadamin and Novikov, Alexander and Balog, Matej and Kumar, M Pawan and Dupont, Emilien and Ruiz, Francisco JR and Ellenberg, Jordan S and Wang, Pengming and Fawzi, Omar and others},
  journal={Nature},
  volume={625},
  number={7995},
  pages={468--475},
  year={2024},
  publisher={Nature Publishing Group UK London}
}

@inproceedings{MemoryR1,
    title = "Memory-R1: Enhancing Large Language Model Agents to Manage and Utilize Memories via Reinforcement Learning",
    author = "Yan, Sikuan  and
      Yang, Xiufeng  and
      Huang, Zuchao  and
      Nie, Ercong  and
      Ding, Zifeng  and
      Li, Zonggen  and
      Ma, Xiaowen  and
      Bi, Jinhe  and
      Kersting, Kristian  and
      Pan, Jeff Z.  and
      Schuetze, Hinrich  and
      Tresp, Volker  and
      Ma, Yunpu",
    booktitle = "Proceedings of the 64th Annual Meeting of the {A}ssociation for {C}omputational {L}inguistics (Volume 1: Long Papers)",
    month = jul,
    year = "2026",
    pages = "12805--12825",
    ISBN = "979-8-89176-390-6",
}

@inproceedings{FunctionsAsLearnableWeights,
  title  = {Offline Training of Language Model Agents with Functions as Learnable Weights},
  author = {Zhang, Shaokun and Zhang, Jieyu and Liu, Jiale and Song, Linxin and Wang, Chi and Krishna, Ranjay and Wu, Qingyun},
  year   = {2024},
  booktitle = {Proceedings of the 41st International Conference on Machine Learning},
  pages = 	 {60315--60335},
  volume = 	 {235},
  series = 	 {Proceedings of Machine Learning Research},
  month = 	 {21--27 Jul},
  publisher =    {PMLR},
}

@inproceedings{ReAct,
    title={ReAct: Synergizing Reasoning and Acting in Language Models},
    author={Shunyu Yao and Jeffrey Zhao and Dian Yu and Nan Du and Izhak Shafran and Karthik R Narasimhan and Yuan Cao},
    booktitle={The Eleventh International Conference on Learning Representations },
    year={2023},
}

@article{rsi_survey1,
	doi = {10.20944/preprints202608.0051.v1},
	year = 2026,
	month = {August},
	publisher = {Preprints},
	author = {Shuaiqi Liu and Zhengkai Lin and Yuxiang Zhang and Yuanyi Ren and Yue Wu and Yongbin Li and Zheng Wang and Zhihang Fu and Jieping Ye},
	title = {The Path to Recursive Self-Improving Agents: Foundation, Framework, and Future Directions},
	journal = {Preprints}
}

@article{rsi_survey2,
  title={Self-Improvements in Modern Agentic Systems: A Survey},
  author={Ren, Zhe and Chen, Yimeng and Guo, Dandan and Rong, Guowei and Li, Tonghui and Xiong, RB and Lan, Qingfeng and Wang, Wenyi and Nanbo, Li and Yang, Yibo and others},
  journal={arXiv preprint arXiv:2607.13104},
  year={2026}
}

@article{rsi_survey3,
  title={The Last AI Built by Humans: Toward Genuine Recursive Self-Improvement},
  author={Duan, Yi and Liu, Ying and Tang, Zirui and Chen, Haodong and Zhou, Jun and Liu, Yumou and Xu, Bangrui and Wu, Yukai and Chen, Sidi and Zhou, Yuhan and others},
  journal={arXiv preprint arXiv:2609.11873},
  year={2026}
}

@inproceedings{reflexion,
    title={Reflexion: language agents with verbal reinforcement learning},
    author={Noah Shinn and Federico Cassano and Ashwin Gopinath and Karthik R Narasimhan and Shunyu Yao},
    booktitle={Thirty-seventh Conference on Neural Information Processing Systems},
    year={2023},
}

@article{SkillClaw,
  title={Skillclaw: Let skills evolve collectively with agentic evolver},
  author={Ma, Ziyu and Yang, Shidong and Ji, Yuxiang and Wang, Xucong and Wang, Yong and Hu, Yiming and Huang, Tongwen and Chu, Xiangxiang},
  journal={arXiv preprint arXiv:2604.08377},
  year={2026}
}

@article{Voyager,
    title={Voyager: An Open-Ended Embodied Agent with Large Language Models},
    author={Guanzhi Wang and Yuqi Xie and Yunfan Jiang and Ajay Mandlekar and Chaowei Xiao and Yuke Zhu and Linxi Fan and Anima Anandkumar},
    journal={Transactions on Machine Learning Research},
    issn={2835-8856},
    year={2024},
}

@inproceedings{AgentSquare,
  title     = {{AgentSquare}: Automatic {LLM} Agent Search in Modular Design Space},
  author    = {Shang, Yu and Li, Yu and Zhao, Keyu and Ma, Likai and Liu, Jiahe and Xu, Fengli and Li, Yong},
  booktitle = {The Thirteenth International Conference on Learning Representations},
  year      = {2025},
}

@inproceedings{reasoningbank,
  title     = {{ReasoningBank}: Scaling Agent Self-Evolving with Reasoning Memory},
  author    = {Ouyang, Siru and Yan, Jun and Hsu, I-Hung and Chen, Yanfei and Jiang, Ke and Wang, Zifeng and Han, Rujun and Le, Long T. and Daruki, Samira and Tang, Xiangru and Tirumalashetty, Vishy and Lee, George and Rofouei, Mahsan and Lin, Hangfei and Han, Jiawei and Lee, Chen-Yu and Pfister, Tomas},
  booktitle = {The Fourteenth International Conference on Learning Representations},
  year      = {2026},
}

@article{AdaptiveAutoHarness,
  title   = {Adaptive Auto-Harness: Sustained Self-Improvement for Agentic System Deployment on Open-Ended Task Streams},
  author  = {Liu, Zewen and Shi, Zhan and Sang, Yisi and He, Bing and Lin, Minhua and Wei, Tianxin and Wang, Dakuo and Dumoulin, Benoit and Jin, Wei and Lu, Hanqing},
  journal = {arXiv preprint arXiv:2606.01770},
  year    = {2026},
}

@article{SelfHarness,
  title   = {{Self-Harness}: Harnesses That Improve Themselves},
  author  = {Zhang, Hangfan and Zhang, Shao and Li, Kangcong and Zhang, Chen and Chen, Yang and Zhang, Yiqun and Bai, Lei and Hu, Shuyue},
  journal = {arXiv preprint arXiv:2606.09498},
  year    = {2026},
}
\bibliographystyle{iclr2027_conference}

\newpage
\appendix
\section{Reproducibility statement}
\label{app:reprod-statement}

We describe the optimization procedure in Section~\ref{sec:method} and Algorithm~\ref{alg:plugharness}, and the evaluation protocol in Section~\ref{sec:experimental-setup}.
Appendix~\ref{app:implmentation_details} reports model configurations, search hyperparameters and the API access period.
Appendices~\ref{app:system-prompt} and~\ref{app:seed-harness} provide the initial harness, seed plugin sources, and proposer prompts.
Appendix~\ref{app:dataset-split} describes the SWE-Bench Verified split, and Appendix~\ref{app:schema} specifies the plugin interfaces. 
An anonymous code repository is linked in the abstract.

\section{Ethics statement}
\label{app:ethic-statement}

The experiments and research within the scope of this paper raise few ethical concerns, such as potentially harmful insights and discrimination/bias/fairness concerns.
Our experiments use existing benchmarks for software engineering, command-line interaction, and question answering.
Automatically evolved harnesses can improve software development, but generated code and commands may also introduce errors or unintended changes. Deployment beyond benchmark environments should therefore include code review and appropriate execution permissions.

\section{Use of AI Statement}
\label{app:llm_statement}

We used generative AI tools to assist with literature review, code development and manuscript polishing.
In particular, GPT-5.6 Sol assisted with mechanism extraction for the seed plugin library (as stated in Appendix~\ref{app:seed-harness}), and all resulting plugins underwent manual inspection.
The use of LLMs as proposers and solvers is described in the method and implementation details (as stated in Appendix~\ref{app:implmentation_details}).

\section{Limitations and Future Work}
\label{app:limitations}
Despite its effectiveness, this work faces several limitations.
First, we evaluate harness evolution on fixed task sets over a limited number of iterations. Longer-term evolution under changing task distributions remains to be explored, together with the management of a growing plugin library.
Second, the transfer of discovered mechanisms depends on the task domain and solver model. Gains are smaller across dissimilar QA domains; thus, optimizing across multiple domains and models could improve broader reuse.
Third, plugin mutation evaluates each change with the remaining harness fixed. This can miss mechanisms that require coordinated changes to other plugins or the workflow.
Future work could explore harness optimization on long-horizon tasks, more significant cross-domain transferability and the plugin-workflow joint optimization.

\section{Implementation Details}
\label{app:implmentation_details}

For the SWE-Bench experiments in Table~\ref{tab:main-results}, we use GPT-5.6 Terra and Kimi-K3 as proposers in the respective optimization settings, with the same backbone serving as the optimization solver.
Harnesses are selected using held-in scores from the corresponding optimization solver and then fixed for evaluation across GPT-5.6 Terra, Kimi-K3, and GLM-5.2.
For Terminal-Bench, QA and all ablation experiments, Kimi-K3 serves as both proposer and optimization solver.
Reasoning effort is set to \texttt{xhigh} for the proposer and \texttt{none} for all solvers.
At each iteration, \plugharness generates $N=8$ plugin mutation candidates per iteration and evaluates each on a separately sampled minibatch of $b=20$ tasks. 

Each minibatch contains 10 tasks solved by the current harness and 10 unsolved tasks.
For example, on SWE-Bench Verified, each iteration evaluates mutation candidates and one recomposition candidate using $|\mathcal D_\mathrm{val}|+N\times b=400+8\times20=560$ solver rollouts.
We run our method for $T=15$ iterations and the baselines for $T=25$ iterations.
The subsequent harness recomposition stage generates one candidate, which is evaluated on all held-in tasks.
Tables~\ref{tab:implementation-hyperparameters} and~\ref{tab:implementation-models} summarize the search hyperparameters and model configurations.
For all models, we use response APIs accessed in September 2026.
It is worth noting that we observe an obvious performance degradation in GPT-series models around the end of August.
Although the cause of this degradation remains unclear, we reran all experiments in September to ensure a fair comparison.

\paragraph{Baseline implementations.}
ReAct follows a reasoning--acting loop with either zero-shot instructions or few-shot demonstrations.
ACE augments the harness with a context playbook that accumulates experience. In our experiment we only turn on the ``online'' memory of ACE with no cross-task experience. Thus, it is considered an ``expert-curated'' baseline instead of an automatically updated one.

All the automatically optimized harnesses start from a ReAct-style initial harness.
GEPA revises prompt instructions using execution feedback, where we let it recursively optimize the solver's system prompt.
DGM and Meta-Harness revise complete executable harness implementations from the initial harness.
All methods use the same task splits, API endpoints, and environment sandboxes, with model backbones following the corresponding experimental setting.
All automatically optimized baselines run for 25 iterations.
The selected harness is then kept fixed for held-out and cross-model evaluation.

\paragraph{Plugin reuse setup.}
The evolved harness and plugin library used in Figure~\ref{fig:analysis-reuse} come from the search that produced the best held-in-selected \plugharness{} result in the corresponding optimization setting of Table~\ref{tab:main-results}.
The three conditions start from the initial harness with the initial library, the same initial harness with the evolved library, and the selected evolved harness with its evolved library, respectively.
The subsequent harness optimization runs on the held-out 100 tasks following the held-in protocol in Section~\ref{sec:experimental-setup}.

\begin{table}[ht]
  \centering
  \caption{Hyperparameters.}
  \vspace{5pt}
  \small
  \label{tab:implementation-hyperparameters}
  \renewcommand{\arraystretch}{1.25}
  \resizebox{0.7\linewidth}{!}{
  \begin{tabular}{l@{\hspace{10em}}c}
    \toprule
    Hyperparameter & Value \\
    \midrule
    \rowcolor{gray!12}
    \multicolumn{2}{c}{\textit{Proposer Agent}} \\
    Optimization iterations ($T$) & 15 \\
    Mutation candidates per iteration ($N$) & 8 \\
    Mutation minibatch size ($b$) & 20 \\
    Recomposition candidates per iteration & 1 \\
    Solved:Unsolved tasks per minibatch & 10:10 \\
    Max interaction turns & 40 \\
    Random Seed & 42 \\
    \midrule
    \rowcolor{gray!12}
    \multicolumn{2}{c}{\textit{Solver Agent}} \\
    Max interaction turns & 20 \\
    Timeout seconds & 360 \\
    \bottomrule
  \end{tabular}
  }
\end{table}

\begin{table}[ht]
  \centering
  \caption{Model configurations for the proposer and solvers.}
  \vspace{5pt}
  \small
  \label{tab:implementation-models}
  \renewcommand{\arraystretch}{1.25}
  \resizebox{0.8\linewidth}{!}{
      \begin{tabular}{l@{\hspace{5em}}l@{\hspace{5em}}l@{\hspace{5em}}r}
            \toprule
            Role & Model & Reasoning & Max Output Tokens \\
            \midrule
            Proposer & GPT-5.6 Terra & \texttt{xhigh} & 16,384 \\
            Proposer & Kimi-K3 & \texttt{xhigh} & 16,384 \\
            Solver & GPT-5.6 Terra & \texttt{none} & 8,192 \\
            Solver & Kimi-K3 & \texttt{none} & 8,192 \\
            Solver & GLM-5.2 & \texttt{disabled} & 8,192 \\
            \bottomrule
      \end{tabular}%
  }
\end{table}

\section{Additional Results}
\label{app:additional-results}

We further compare against SkillOPT~\citep{skillopt}, A-Evolve~\citep{a-evolve} and
Agentic Harness Engineering (AHE)~\citep{AgenticHarnessEngineering} on SWE-bench Verified with Kimi-K3 as both proposer and solver.

The results in Table~\ref{tab:app-swe} show that the advantage of \plugharness{} also holds over the additional baselines.
Compared with SkillOPT, A-Evolve, and AHE, \plugharness{} improves held-out resolve rates by 9.00, 12.00, and 6.00 points, respectively.
These results further support the effectiveness of plugin-oriented harness optimization beyond the tasks used during search.
For computational burdens, we do not put these baseline methods in \S~\ref{sec:main-results} for comparison.

\begin{table}
  \centering
  \caption{
    Results on SWE-bench Verified with Kimi-K3 as both proposer and solver.
    Values are resolve rates (\%).
    Bold marks the best result in each column.
  }
  \label{tab:app-swe}
  \vspace{7pt}
  \small
  \setlength{\tabcolsep}{4pt}
  \renewcommand{\arraystretch}{1.05}
  \resizebox{0.8\linewidth}{!}{
    \begin{tabular}{l@{\hspace{5em}}c@{\hspace{5em}}c}
      \toprule
      Method & Held-In (400) & Held-Out (100) \\
      \midrule
      GEPA             & 61.00 & 62.00 \\
      DGM              & 58.00& 55.00 \\
      SkillOPT~\citep{skillopt}              & 54.75 & 60.00 \\
      A-Evolve~\citep{a-evolve}              & 61.75 & 57.00 \\
      AHE~\citep{AgenticHarnessEngineering}              & 60.25 & 63.00 \\
      Meta-Harness     & 64.25 & 63.00\\
      \textbf{\plugharness} & \textbf{66.75} & \textbf{69.00} \\
      \bottomrule
    \end{tabular}
  }
\end{table}

\section{Initial Harness}
\label{app:seed-harness}
\textbf{Seed Plugin Library.}
We construct the seed plugin library $\mathcal{L}_0$ from prior research and open-source agent systems, with sources summarized in Table~\ref{tab:seed-plugin-sources}. We manually curate the paper list and use GPT-5.6 Sol to extract role responsibilities and prompts, reusable skill procedures, tool schemas, and memory record structures. We adapt these assets into \texttt{Python} plugins with standardized interfaces and manually inspect all plugins. Each plugin retains its configuration, supporting resources, and provenance documenting its sources and adaptation. The resulting library contains 75 components across four categories (Table~\ref{tab:seed-plugin-components}), from which each harness selects the components it uses.

Roles render prompts specifying responsibilities and constraints, with input/output metadata where applicable. Tools expose structured argument schemas and executable operations: repository actions run in the task environment, while patch submission writes a prediction artifact. Skills provide loadable instructions and supporting resources. Memory components manage the mechanisms to store and retrieve task-local evidence using lexical overlap.

\begin{table}[t]
    \centering
    \caption{Sources used to construct the seed plugin library. Repository and specification names link to their corresponding resources.}
    \vspace{5pt}
    \label{tab:seed-plugin-sources}
    \small
    \setlength{\tabcolsep}{5pt}
    \renewcommand{\arraystretch}{1.15}
    \begin{tabularx}{0.9\linewidth}{@{}>{\raggedright\arraybackslash}p{0.24\linewidth} >{\raggedright\arraybackslash}X@{}}
        \toprule
        \textbf{Source category} & \textbf{Papers and resources} \\
        \midrule
        Research papers &
        SWE-agent~\citep{SWEAgent}, Agentless~\citep{agentless}, SWE-smith~\citep{swesmith}, and SWE-Bench~\citep{SWEBench}. \\
        \addlinespace[4pt]
        Agent implementations &
        \href{https://github.com/SWE-agent/mini-swe-agent}{mini-SWE-agent},
        \href{https://github.com/Aider-AI/aider}{Aider},
        \href{https://github.com/deepseek-ai/deepseek-harness}{DeepSeek Harness}, and
        \href{https://github.com/OpenHands/software-agent-sdk}{OpenHands Software Agent SDK}. \\
        \addlinespace[4pt]
        Skill instructions and organization &
        \href{https://github.com/agentskills/agentskills}{Agent Skills specification},
        \href{https://github.com/anthropics/skills}{Anthropic Skills},
        \href{https://github.com/obra/superpowers}{Superpowers},
        \href{https://github.com/addyosmani/agent-skills}{Addy Osmani Agent Skills}, and
        \href{https://github.com/OpenHands/extensions}{OpenHands Extensions}. \\
        \addlinespace[4pt]
        Trajectories and evaluation records &
        \href{https://github.com/SWE-Bench/experiments}{SWE-Bench experiment repository} and
        \href{https://huggingface.co/datasets/SWE-Bench/SWE-smith-trajectories}{SWE-smith trajectory corpus}. \\
        \bottomrule
    \end{tabularx}
\end{table}

\begin{table}[t]
    \centering
    \caption{Composition of the seed plugin library. Coverage is summarized by function.}
    \vspace{5pt}
    \label{tab:seed-plugin-components}
    \small
    \setlength{\tabcolsep}{5pt}
    \renewcommand{\arraystretch}{1.15}
    \begin{tabularx}{0.9\linewidth}{@{}l r >{\raggedright\arraybackslash}X@{}}
        \toprule
        \textbf{Category} & \textbf{Count} & \textbf{Functional coverage} \\
        \midrule
        Roles & 20 &
        Baseline coding and boundary enforcement; task analysis and context preparation; fault localization and diagnosis; patch planning, generation, and ranking; verification and finalization; experience extraction and plugin or workflow editing. \\
        \addlinespace[5pt]
        Tools & 16 &
        Terminal and shell execution; file access and repository navigation; text and symbol search; patch application; Git inspection; test discovery and execution; linting and failure inspection; patch submission. \\
        \addlinespace[5pt]
        Skills & 31 &
        Task analysis and repository orientation; skill routing and instruction loading; code localization and call-chain tracing; reproduction and debugging; patching and contract preservation; testing and completion checks; blocker diagnosis and recovery; lesson extraction and skill improvement; harness-authoring guidance. \\
        \addlinespace[5pt]
        Memory & 8 &
        A baseline experience bank and seven structured record types: trajectories, task outcomes, failure signatures, lesson candidates, counterexamples, component credit, and skill evaluations. \\
        \midrule
        \textbf{Total} & \textbf{75} & \\
        \bottomrule
    \end{tabularx}
\end{table}

\begin{figure}[t]
    \centering
    \begin{tcolorbox}[
        enhanced,
        width=\linewidth,
        colback=white,
        boxrule=0.5pt,
        boxsep=0pt,
        left=0pt,
        right=0pt,
        top=0pt,
        bottom=0pt,
        sharp corners,
        drop shadow
    ]
        \includegraphics[width=\linewidth]{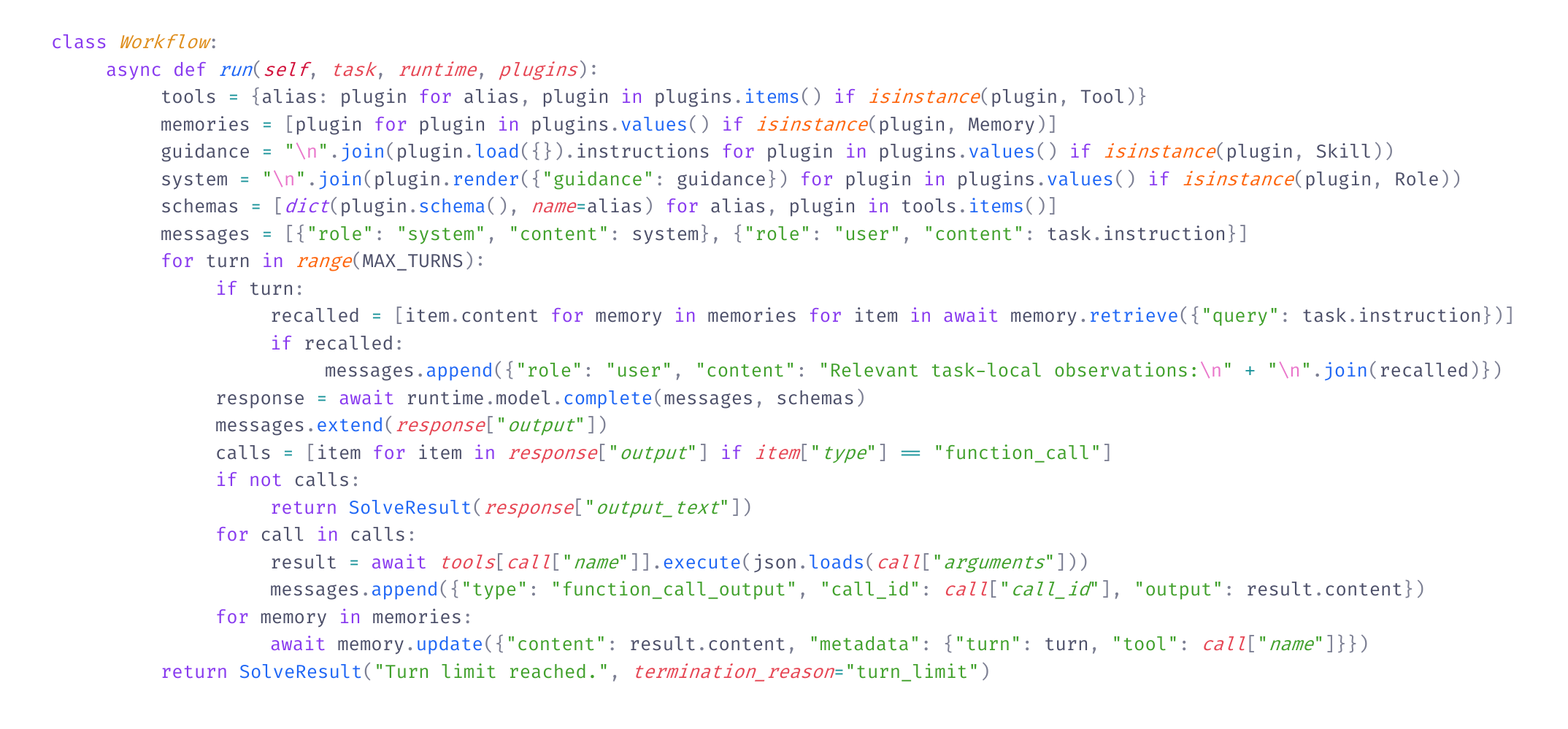}
    \end{tcolorbox}
    \caption{\texttt{Python} implementation of the initial harness workflow.}
    \label{fig:app-seed-harness}
\end{figure}

\textbf{Seed Workflow.}
The initial harness instantiates four plugins from $\mathcal L_0$.
The harness starts with a coder role, a terminal tool, a debugging skill, and an experience-bank memory.
The workflow loads the skill instructions into the role prompt and exposes the selected tools to the solver.
At each turn, it retrieves task-local observations, queries the solver, executes any requested tools, and stores their outputs in memory.
Execution ends when the solver returns without tool calls or reaches the limit of 20 turns.
Memory is initialized separately for each task.
The complete workflow and its plugin configuration are shown in Figure~\ref{fig:app-seed-harness}.

\section{System Prompt for Proposer}
\label{app:system-prompt}

The proposer uses separate instructions for plugin mutation and harness recomposition.
Each proposal starts a new conversation, with the current harness, accessible plugin references, evaluation feedback, and historical candidate summaries supplied as context.
Trajectory summaries provide an overview of the feedback, while file-reading actions allow the proposer to inspect full implementations and relevant trajectory segments.
The phase-specific instructions are given below.

\textbf{Plugin Mutation.}
Each mutation branch receives the current harness and its scores and execution trajectories on a separately sampled minibatch.
The proposer uses this feedback to select a plugin used by the harness and develop a modified implementation, while keeping the workflow, the remaining plugins, and the selected instance's configuration fixed.
The mutation instructions are summarized in \texttt{Plugin Mutation Instructions}.

\textbf{Harness Recomposition.}
The proposer receives the incumbent's full held-in feedback and the mutation results from the current iteration.
It selects plugins from the updated library and revises the workflow to coordinate them.
Plugin implementations remain fixed during this stage.
The detailed raw prompt is included in \texttt{System Prompt for Harness Recomposition}.

Both prompts are followed by a shared file-action protocol, which is included in \texttt{File-Action Protocol}.
The available actions are \texttt{list\_files}, \texttt{read\_file}, \texttt{read\_file\_range}, \texttt{write\_file}, and \texttt{submit\_proposal}.

\begin{promptbox}{Plugin Mutation Instructions}
Use this local experiment to invent a plugin-level solution to a concrete failure in the supplied trajectories. Read summaries, then inspect relevant full trace ranges and plugin code to identify where the attempt went wrong; check successes for regressions.

Inspect the current harness and the supplied minibatch feedback. Develop one transferable mechanism that could fix that failure. You may revise the existing implementation or replace it from scratch; the copied code is only a scaffold. You may revise the selected plugin's implementation or replace it from scratch.

Do not implement explicit content hashing or digests. Use direct string equality for deduplication.

Write the modified implementation to a new \texttt{new\_plugins/<kind>/<name>/<version>/} package for the selected plugin. \texttt{library/} is READ-ONLY. Never write plugins under \texttt{child/plugins/} or any other \texttt{child/} subdirectory.

Keep every workflow file byte-for-byte unchanged. In \texttt{harness.yaml}, change only the selected instance's plugin reference to the new version; keep its alias, configuration, and all other manifest values unchanged.

Create exactly one new plugin with the same kind/name as the selected source reference and a new version. Implement a transferable improvement, not task-specific answers.

Other plugins and the shared runtime/evaluator are immutable. Keep the public interface and existing instance configuration compatible. Do not install dependencies or inspect hidden tests.

Use local feedback and your previous attempts, including failures, to refine the mechanism. In hypothesis, cite the task/trace evidence, the failure mechanism and the expected repair.

Deliver actual files and call \texttt{submit\_proposal} with \texttt{hypothesis}, \texttt{changes:list[string]}, \texttt{new\_plugin\_refs}. New plugin provenance requires \texttt{label}, \texttt{sources}, \texttt{upstream\_assets}, \texttt{adaptation} and \texttt{parent\_refs}.
\end{promptbox}

\begin{promptbox}{System Prompt for Harness Recomposition}
Build a better complete Harness by integrating useful ideas from this round's plugin experiments. Prioritize newly published plugins with positive local score gains; read their implementations and relevant parent/child trajectories to understand which failures they repaired, any regressions, and when their mechanisms should help.

Local gains are evidence for integration, not a guarantee of full-set improvement.

Redesign workflow control flow, context and plugin composition so selected plugins are actually used where they can help. Combine compatible mechanisms; avoid blindly stacking plugins or preserving the incumbent assembly by default. Use rejected attempts to avoid repeating failures.

In hypothesis, name the selected plugin refs, supporting local results and how the workflow will use them; explain if no new plugin is suitable.

Do not implement explicit content hashing or digests. Use direct string equality for deduplication.

You may rewrite the entire child workflow package, helpers, resources and plugin assembly. Use only frozen \texttt{allowed\_plugin\_refs}. Do not create or modify plugin packages in this phase.

Do not change the model, runtime, evaluator, resource ceilings or hidden tests; do not install dependencies. Do not hard-code task answers.

Deliver executable files and \texttt{submit\_proposal} with \texttt{new\_plugin\_refs=[]}.
\end{promptbox}

\begin{promptbox}{File-Action Protocol (shared for both stages).}
Use the same confined file actions described below, but encode them in ordinary text.

Every response must be one JSON object:
\par\smallskip
{\ttfamily\footnotesize\raggedright
\setlength{\parindent}{0pt}
\noindent\{\\
\hspace*{0.8em}"actions": [\\
\hspace*{1.6em}\{\\
\hspace*{2.4em}"name": "read\_file",\\
\hspace*{2.4em}"arguments": \{\\
\hspace*{3.2em}"path": "child/workflow.py"\\
\hspace*{2.4em}\}\\
\hspace*{1.6em}\}\\
\hspace*{0.8em}],\\
\hspace*{0.8em}"final": false\\
\}\par}
\smallskip

You may batch actions. Their results arrive as quoted data in the next user message.

Use \texttt{final=true} only after delivering all executable files and \texttt{proposal\_result.json}. Use \texttt{submit\_proposal} for final metadata. Each changes entry must be a string, not an object.

Never emit API function calls, shell commands as actions, prose outside JSON, or a mere diff description.

File actions have exactly the same read/write scope as the supplied coding-agent contract.
\end{promptbox}

\section{Dataset Split}
\label{app:dataset-split}

\textbf{SWE-Bench Verified.}
We use a fixed partition of the 500 tasks in SWE-Bench Verified into a held-in split $\mathcal{D}_{\mathrm{val}}$ of 400 tasks and a held-out split $\mathcal{D}_{\mathrm{test}}$ of 100 tasks. The partition is specified by disjoint task-ID lists and is shared across all methods and solver backbones. During optimization, mutation minibatches are sampled exclusively from $\mathcal{D}_{\mathrm{val}}$, and recomposition candidates are evaluated on the full held-in split.
We select the harness with the highest held-in resolve rate under the corresponding optimization solver and keep its workflow and plugin configuration fixed for held-out and cross-model evaluation.

\begin{figure}[t]
    \centering
    \includegraphics[width=0.9\linewidth]{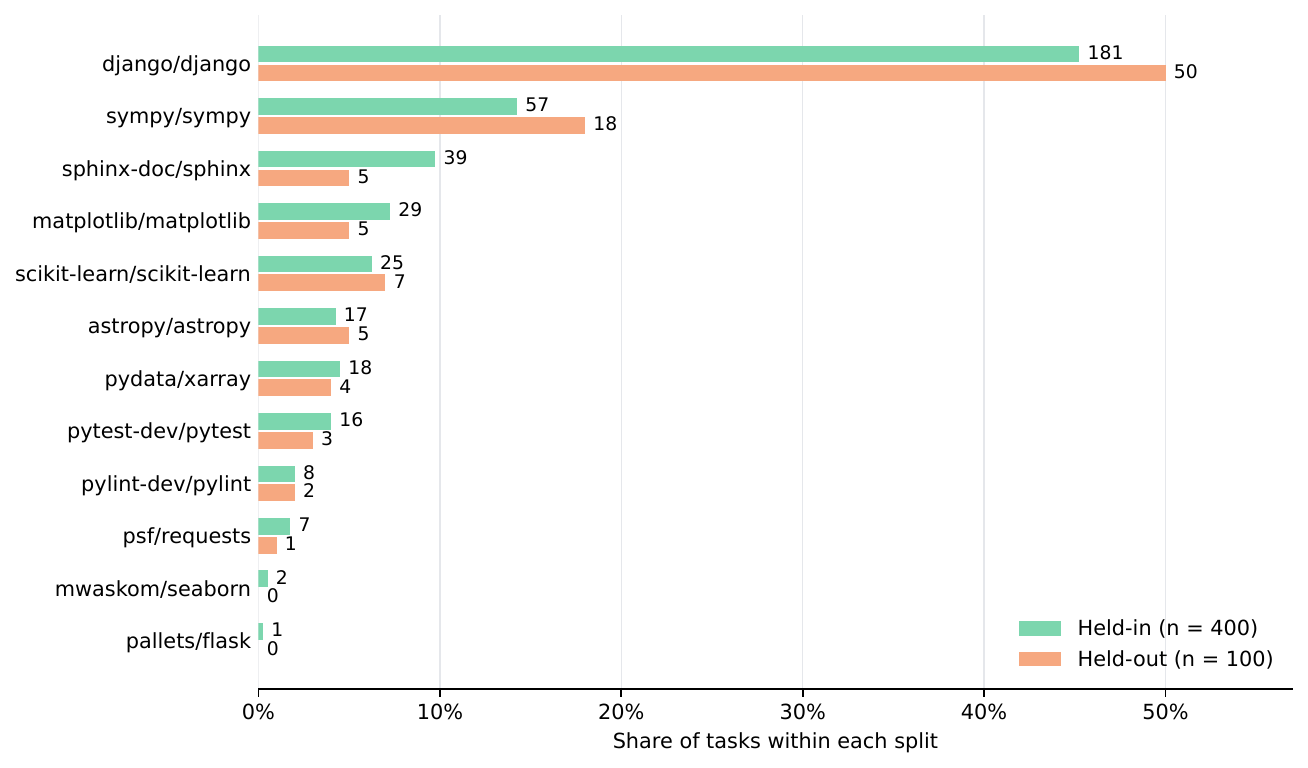}
    \vspace{-5pt}
    \caption{Sub-domain distribution of held-in and held-out tasks on Swe-Bench-Verified.}
    \label{fig:app-subdomain-split}
\end{figure}

\paragraph{Subdomain composition.}
We treat each source repository as a subdomain and compare its proportion within each split in Figure~\ref{fig:app-subdomain-split}. Normalizing by split size allows a direct comparison despite the 4:1 size ratio. Django accounts for 45.25\% of held-in tasks and 50\% of held-out tasks; SymPy accounts for 14.25\% and 18\%, respectively, while Sphinx accounts for 9.75\% and 5\%. The held-in split covers all 12 repositories, whereas the held-out split covers 10: the two Seaborn tasks and the single Flask task occur only in the held-in split. Thus, the held-out evaluation measures task-level generalization within the benchmark, rather than transfer to unseen repositories, and does not provide held-out evidence for Seaborn or Flask.

\section{Plugin Schema}
\label{app:schema}

\begin{figure*}[t]
  \centering
  \begin{minipage}[t]{0.48\textwidth}
    \centering
    \textbf{(a) Plugin manifest: \texttt{plugin.yaml}}\par
    \vspace{0pt}
    \begin{tcolorbox}[
        enhanced,
        width=\linewidth,
        colback=white,
        boxrule=0.5pt,
        boxsep=0pt,
        left=0pt,
        right=0pt,
        top=0pt,
        bottom=0pt,
        sharp corners,
        drop shadow
    ]
        \includegraphics[width=\linewidth]{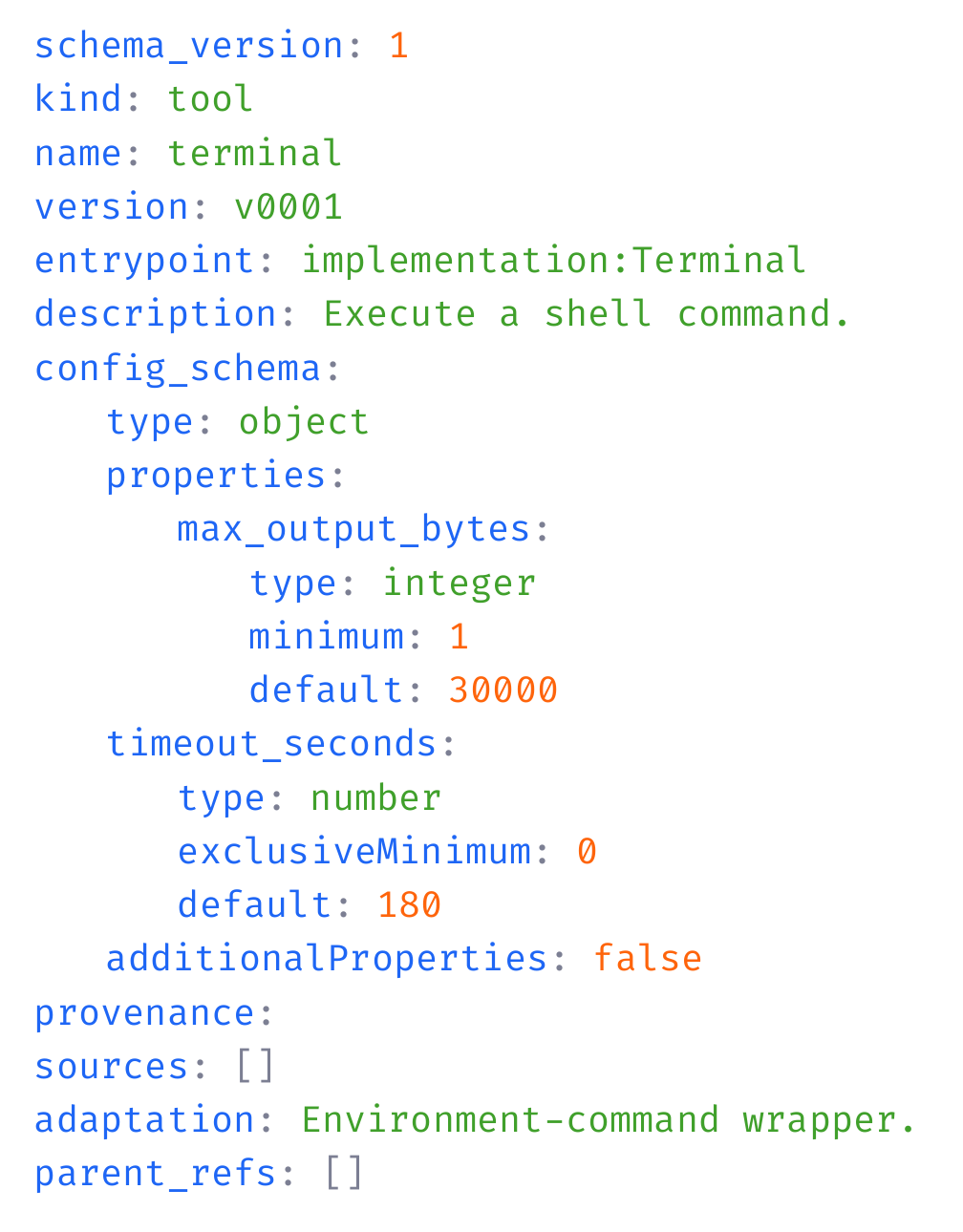}
    \end{tcolorbox}
  \end{minipage}\hfill
  \begin{minipage}[t]{0.48\textwidth}
    \centering
    \textbf{(b) Harness manifest: \texttt{harness.yaml}}\par
    \vspace{0pt}
    \begin{tcolorbox}[
        enhanced,
        width=\linewidth,
        colback=white,
        boxrule=0.5pt,
        boxsep=0pt,
        left=0pt,
        right=0pt,
        top=0pt,
        bottom=0pt,
        sharp corners,
        drop shadow
    ]
        \includegraphics[width=\linewidth]{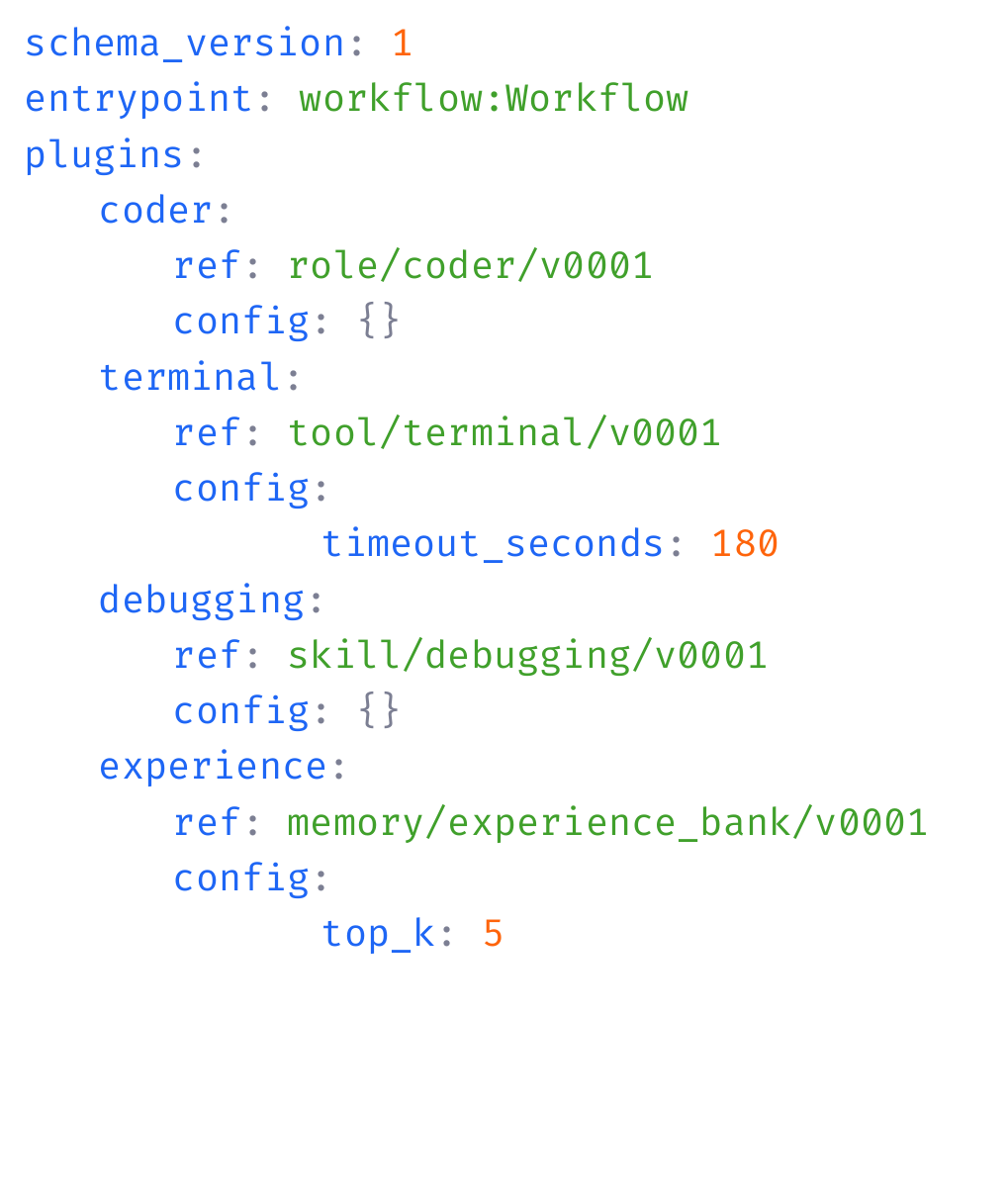}
    \end{tcolorbox}
  \end{minipage}
  \caption{Examples of plugin and harness manifests. (a) \texttt{plugin.yaml} defines a terminal tool and its instance configuration. (b) \texttt{harness.yaml} assembles a harness by mapping workflow aliases to versioned plugins and their configurations.}
  \label{fig:plugin-schema}
\end{figure*}

\paragraph{Plugin packages.}
Each plugin is implemented as a versioned Python package stored at \texttt{<library>/<kind>/<name>/<version>/}. The package contains a \texttt{plugin.yaml} manifest, \texttt{\_\_init\_\_.py}, an implementation module, and supporting resources such as prompts and rubrics. The manifest specifies how the plugin is loaded and configured, while the Python implementation defines its behavior. Figure~\ref{fig:plugin-schema} shows an example plugin manifest and its use in a harness.

\paragraph{Plugin manifest.}
All four plugin categories share the same manifest structure. The fields \texttt{kind}, \texttt{name}, and \texttt{version} identify a plugin through a reference such as \texttt{tool/terminal/v0001}, and \texttt{entrypoint} specifies its implementation class. Instance configurations are defined by \texttt{config\_schema} using JSON Schema. The loader fills in default values and validates the configuration before creating the instance. For example, the terminal plugin in Figure~\ref{fig:plugin-schema}(a) specifies an output limit and an execution timeout. Optional \texttt{provenance} metadata records the plugin's sources, adaptation details, and parent references.

\paragraph{Plugin interfaces.}
Each plugin implements the Python interface of its category, as summarized in Table~\ref{tab:plugin-interfaces}. The workflow invokes these methods to render role prompts, load skill instructions, execute tools, and manage memory. Tool-call argument schemas are exposed through \texttt{schema()}, separately from the instance configuration defined in \texttt{plugin.yaml}.

\begin{table}[t]
  \centering
  \caption{Plugin interfaces exposed to the workflow.}
  \vspace{5pt}
  \label{tab:plugin-interfaces}
  \small
  \renewcommand{\arraystretch}{1.15}
  \begin{tabular}{@{}lp{0.30\linewidth}p{0.48\linewidth}@{}}
    \toprule
    Category & Interface & Function \\
    \midrule
    Role
      & \texttt{render(context)}
      & Returns a prompt string constructed from the supplied context. \\
    Tool
      & \texttt{schema()}\newline \texttt{execute(arguments)}
      & Exposes the tool-call schema and asynchronously executes an operation, returning a \texttt{ToolResult}. \\
    Skill
      & \texttt{load(context)}
      & Returns a \texttt{SkillContent} containing instructions and associated resources. \\
    Memory
      & \texttt{update(experience)}\newline \texttt{retrieve(query)}
      & Asynchronously stores an experience and retrieves a list of \texttt{MemoryItem} objects. \\
    \bottomrule
  \end{tabular}
\end{table}

\paragraph{Harness assembly.}
The \texttt{harness.yaml} manifest selects plugin versions and assigns each instance an alias and configuration. The loader checks the selected packages and their interfaces, then instantiates them with the specified configurations. The workflow accesses these instances through the \texttt{plugins} mapping; for example, \texttt{plugins["terminal"]} refers to the terminal instance in Figure~\ref{fig:plugin-schema}(b). Its asynchronous \texttt{run(task, runtime, plugins)} method coordinates model calls and plugin execution and returns a \texttt{SolveResult}. Each versioned plugin reference identifies a fixed implementation. Plugin mutation creates a new version and updates the corresponding reference while preserving the instance alias, configuration, and category interface. Harness recomposition selects existing plugin versions and revises the workflow that coordinates them.



\definecolor{dpBorder}{HTML}{4D4D4D}
\definecolor{dpShadow}{HTML}{BFBFBF}
\definecolor{dpText}{HTML}{252525}
\definecolor{dpBlue}{HTML}{2468FF}
\definecolor{dpGreen}{HTML}{339B22}
\definecolor{dpOrange}{HTML}{F26815}
\definecolor{dpComment}{HTML}{858598}

\lstdefinestyle{discoveredpython}{
  language=Python,
  basicstyle=\ttfamily\fontsize{8}{10}\selectfont\color{dpText},
  keywordstyle=\color{dpBlue},
  stringstyle=\color{dpGreen},
  commentstyle=\color{dpComment},
  morekeywords={async,await},
  showstringspaces=false,
  keepspaces=true,
  columns=fullflexible,
  breaklines=true,
  breakatwhitespace=false,
  breakindent=8pt,
  tabsize=4,
  numbers=none,
  aboveskip=3pt,
  belowskip=0pt,
  xleftmargin=0pt,
  literate=
    {0}{{{\color{dpOrange}0}}}1
    {1}{{{\color{dpOrange}1}}}1
    {2}{{{\color{dpOrange}2}}}1
    {3}{{{\color{dpOrange}3}}}1
    {4}{{{\color{dpOrange}4}}}1
    {5}{{{\color{dpOrange}5}}}1
    {6}{{{\color{dpOrange}6}}}1
    {7}{{{\color{dpOrange}7}}}1
    {8}{{{\color{dpOrange}8}}}1
    {9}{{{\color{dpOrange}9}}}1,
  moredelim=**[is][\color{dpOrange}\bfseries]{|!}{!|}
}

\newtcolorbox{discoveredpanel}[1]{
  enhanced,
  colback=white,
  colframe=dpBorder,
  boxrule=0.45pt,
  sharp corners,
  boxsep=0pt,
  left=8pt,
  right=8pt,
  top=8pt,
  bottom=8pt,
  before skip=0pt,
  after skip=0pt,
  equal height group=#1,
  shadow={3pt}{-3pt}{0pt}{fill=dpShadow,opacity=1}
}

\newcommand{\dppaneltitle}[1]{%
  {\centering\bfseries\normalsize #1\par}%
  \vspace{6pt}%
}

\newcommand{\dplabel}[1]{%
  \par\noindent{\footnotesize\bfseries #1}%
  \par\nobreak\vspace{4pt}%
}

\section{Discovered Plugins}
\label{app:case_plugins}

We examine how plugin mutation changes the mechanisms that guide solver behavior. Table~\ref{tab:discovered-plugins} summarizes eight selected mutations across roles, skills, memory, and tools, with one example from each category illustrated below.

\begin{table}[t]
  \centering
  \caption{Selected plugin mutations and their screening results. Values are resolved tasks before and after mutation on each candidate's minibatch, using cached outcomes for the original harness. For incremental evidence retrieval, one verifier timeout leaves 19 tasks with valid outcomes in both evaluations.}
  \label{tab:discovered-plugins}
  \small
  \renewcommand{\arraystretch}{1.12}
  \setlength{\tabcolsep}{4pt}
  \begin{tabularx}{\linewidth}{@{}lXcc@{}}
    \toprule
    Category & Discovered mechanism & Before & After \\
    \midrule
    Role & Repair-scope control & 10/20 & 13/20 \\
         & Discriminative checks and lifecycle continuity & 10/20 & 13/20 \\
    \addlinespace[3pt]
    Skill & Semantic source preservation & 10/20 & 13/20 \\
          & Explicit precedence and representation checks & 10/20 & 13/20 \\
          & Public-path checks with reachability controls & 10/20 & 14/20 \\
    \addlinespace[3pt]
    Memory & Incremental evidence retrieval & 10/19 & 14/19 \\
           & Test-outcome classification & 10/20 & 12/20 \\
    \addlinespace[3pt]
    Tool & Failure-aware terminal feedback & 10/20 & 14/20 \\
    \bottomrule
  \end{tabularx}
\end{table}


\paragraph{Role: repair-scope control.}
The mutated coder role refines how the solver determines the scope of a repair. While the original role encourages small edits, the mutated instructions place guards, transactions, or cleanup around the smallest sequence of persistent mutations that requires them (Figure~\ref{fig:discovered-role}). In \texttt{django\_\_django-16100}, the original solver wraps the entire admin view in a transaction, whereas the solver using the mutated role limits the transaction to the affected bulk-write loop and resolves the task. This case illustrates how a general preference for small edits becomes an explicit rule for locating the repair boundary.

\begin{figure}[t]
  \centering
  \begin{minipage}[t]{0.475\linewidth}
    \vspace{0pt}
    \dppaneltitle{(a) Before mutation}
    \begin{discoveredpanel}{dp-role}
      \dplabel{Role guidance: \texttt{system\_prompt.md}}
      {\footnotesize\raggedright
      Work in short evidence loops: orient and locate the responsible code, obtain a minimal reproduction or identify the closest existing test, then make the smallest plausible edit and add or update a focused regression test.\par}
    \end{discoveredpanel}
  \end{minipage}\hfill%
  \begin{minipage}[t]{0.475\linewidth}
    \vspace{0pt}
    \dppaneltitle{(b) After mutation}
    \begin{discoveredpanel}{dp-role}
      \dplabel{Role guidance: \texttt{system\_prompt.md}}
      {\footnotesize\raggedright
      For a partial batch write, validation failure, transaction, cache update, or state reset, put the guard, rollback scope, or cleanup at the \textbf{smallest sequence of persistent mutations that needs it}.\par}
    \end{discoveredpanel}
  \end{minipage}
  \vspace{4pt}
  \caption{Repair-scope control in the coder role. The mutated instructions specify where to place guards, transactions, or cleanup when repairing persistent mutations.}
  \label{fig:discovered-role}
\end{figure}


\paragraph{Skill: semantic source preservation.}
The debugging skill adds explicit checks on where a repaired operation should obtain its values. The mutated instructions require the solver to identify the intended source and define the behavior when it is absent (Figure~\ref{fig:discovered-skill}). In \texttt{pydata\_\_xarray-6461}, the original solver falls back to the condition's attributes when the source is scalar. With the mutated skill, it returns empty attributes and resolves the task. The change preserves the intended source semantics rather than introducing a fallback solely to avoid an exception.

\begin{figure}[t]
  \centering
  \begin{minipage}[t]{0.475\linewidth}
    \vspace{0pt}
    \dppaneltitle{(a) Before mutation}
    \begin{discoveredpanel}{dp-skill}
      \dplabel{Skill guidance: \texttt{SKILL.md}}
      {\footnotesize\raggedright
      Make the smallest implementation change that enforces the stated invariant; preserve unrelated APIs, ordering, types, and independently meaningful occurrences.\par}
    \end{discoveredpanel}
  \end{minipage}\hfill%
  \begin{minipage}[t]{0.475\linewidth}
    \vspace{0pt}
    \dppaneltitle{(b) After mutation}
    \begin{discoveredpanel}{dp-skill}
      \dplabel{Skill guidance: \texttt{SKILL.md}}
      {\footnotesize\raggedright
      Do not replace an invalid positional access with a fallback to a different input merely because it avoids the exception; decide whether the correct absent-source result is \textbf{empty, a documented default, or an explicitly defined alternative}.\par}
    \end{discoveredpanel}
  \end{minipage}
  \vspace{4pt}
  \caption{Semantic source preservation in the debugging skill. The mutated instructions require an explicit decision about the correct behavior when the intended source is absent.}
  \label{fig:discovered-skill}
\end{figure}


\paragraph{Memory: incremental evidence retrieval.}
The memory plugin changes how execution evidence is returned to the solver (Figure~\ref{fig:discovered-memory}). The original implementation retrieves raw tool outputs by lexical overlap, which can repeatedly return information already present in the conversation. The mutated plugin deduplicates stored outputs and summarizes recent entries that it has not previously returned, together with guidance for subsequent verification. Retrieval thus emphasizes newly collected evidence while reducing repeated memory output.

\begin{figure}[t]
  \centering
  \begin{minipage}[t]{0.475\linewidth}
    \vspace{0pt}
    \dppaneltitle{(a) Before mutation}
    \begin{discoveredpanel}{dp-memory}
      \dplabel{Memory: \texttt{implementation.py}}
\begin{lstlisting}[style=discoveredpython]
# Store every incoming output.
self.entries.append(item)

# Rank outputs by query overlap.
words = set(re.findall(
    r"\w+", str(query).lower()
))
ranked = sorted(
    enumerate(self.entries),
    key=lambda pair: (
        len(words & set(re.findall(
            r"\w+",
            pair[1]["content"].lower()
        ))),
        pair[0]
    ),
    reverse=True
)

# Return the highest-ranked outputs.
return [
    MemoryItem(**item)
    for _, item in
    ranked[:self.config["top_k"]]
]
\end{lstlisting}
    \end{discoveredpanel}
  \end{minipage}\hfill%
  \begin{minipage}[t]{0.475\linewidth}
    \vspace{0pt}
    \dppaneltitle{(b) After mutation}
    \begin{discoveredpanel}{dp-memory}
      \dplabel{Memory: \texttt{implementation.py}}
\begin{lstlisting}[style=discoveredpython]
# Skip duplicate outputs in update().
if any(
    existing["content"] == content
    for existing in self.entries
):
    return

# Retrieve only undelivered entries.
unseen = self.entries[
    |!self._delivered_count:!|
]
self._delivered_count = len(self.entries)
if not unseen:
    return []

# Summarize recent unseen evidence.
selected = unseen[-top_k:]
classified = [
    self._summary(item)
    for item in selected
]
kinds = [
    kind for kind, _ in classified
]
# ... build checkpoint and guidance
\end{lstlisting}
    \end{discoveredpanel}
  \end{minipage}
  \vspace{4pt}
  \caption{Incremental evidence retrieval in the memory plugin. Mutation replaces retrieval by lexical overlap with deduplication and summaries of entries not previously returned. Code excerpts are reformatted and omit supporting logic.}
  \label{fig:discovered-memory}
\end{figure}


\paragraph{Tool: failure-aware terminal feedback.}
The terminal tool adds interpretation of command output to its original truncation behavior (Figure~\ref{fig:discovered-tool}). It inspects the complete output for execution and test failures, including those masked by a successful shell status, and appends recovery guidance. It also tracks the focused test command and its first failure diagnostic, prompting the solver to inspect the failure when the same pair recurs. On its screening minibatch, this mutation resolves five previously unsolved tasks with one regression.

\begin{figure}[t]
  \centering
  \begin{minipage}[t]{0.475\linewidth}
    \vspace{0pt}
    \dppaneltitle{(a) Before mutation}
    \begin{discoveredpanel}{dp-tool}
      \dplabel{Tool: \texttt{implementation.py}}
\begin{lstlisting}[style=discoveredpython]
# Execute the requested command.
result = await (
    self.services.environment.execute(
        command,
        self.config.get(
            "timeout_seconds", 180
        )
    )
)

# Convert the output to text.
content = (
    result.content
    if isinstance(result.content, str)
    else str(result.content)
)

# Bound the output and return it.
result.content = self._truncate(content)
return result
\end{lstlisting}
    \end{discoveredpanel}
  \end{minipage}\hfill%
  \begin{minipage}[t]{0.475\linewidth}
    \vspace{0pt}
    \dppaneltitle{(b) After mutation}
    \begin{discoveredpanel}{dp-tool}
      \dplabel{Tool: \texttt{implementation.py}}
\begin{lstlisting}[style=discoveredpython]
# Interpret output before truncation.
# ... compute status/failure indicators
notice = self._evidence_notice(
    command, raw_content,
    reported_nonzero, explicit_failure
)
notice += self._repeat_failure_notice(
    command, raw_content,
    reported_nonzero, explicit_failure
)
result.content = self._truncate(
    |!raw_content + notice!|
)
return result

# Inside _repeat_failure_notice():
# ... extract the failure diagnostic
focus = self._focused_test_command(command)
previous = self._read_failure_checkpoint()
repeated = (
    previous.get("focus") == focus
    and previous.get("diagnostic")
        == diagnostic
)
self._write_failure_checkpoint(
    focus, diagnostic
)
# ... construct repeated-failure guidance
\end{lstlisting}
    \end{discoveredpanel}
  \end{minipage}
  \vspace{4pt}
  \caption{Failure-aware terminal feedback. The mutated tool interprets command output and tracks repeated failures to guide subsequent checks. Code excerpts are reformatted and omit supporting logic.}
  \label{fig:discovered-tool}
\end{figure}

\FloatBarrier

\section{Token Cost Analysis}
\label{app:token-cost}

\begin{figure}[ht]
    \centering
    \includegraphics[width=\linewidth]{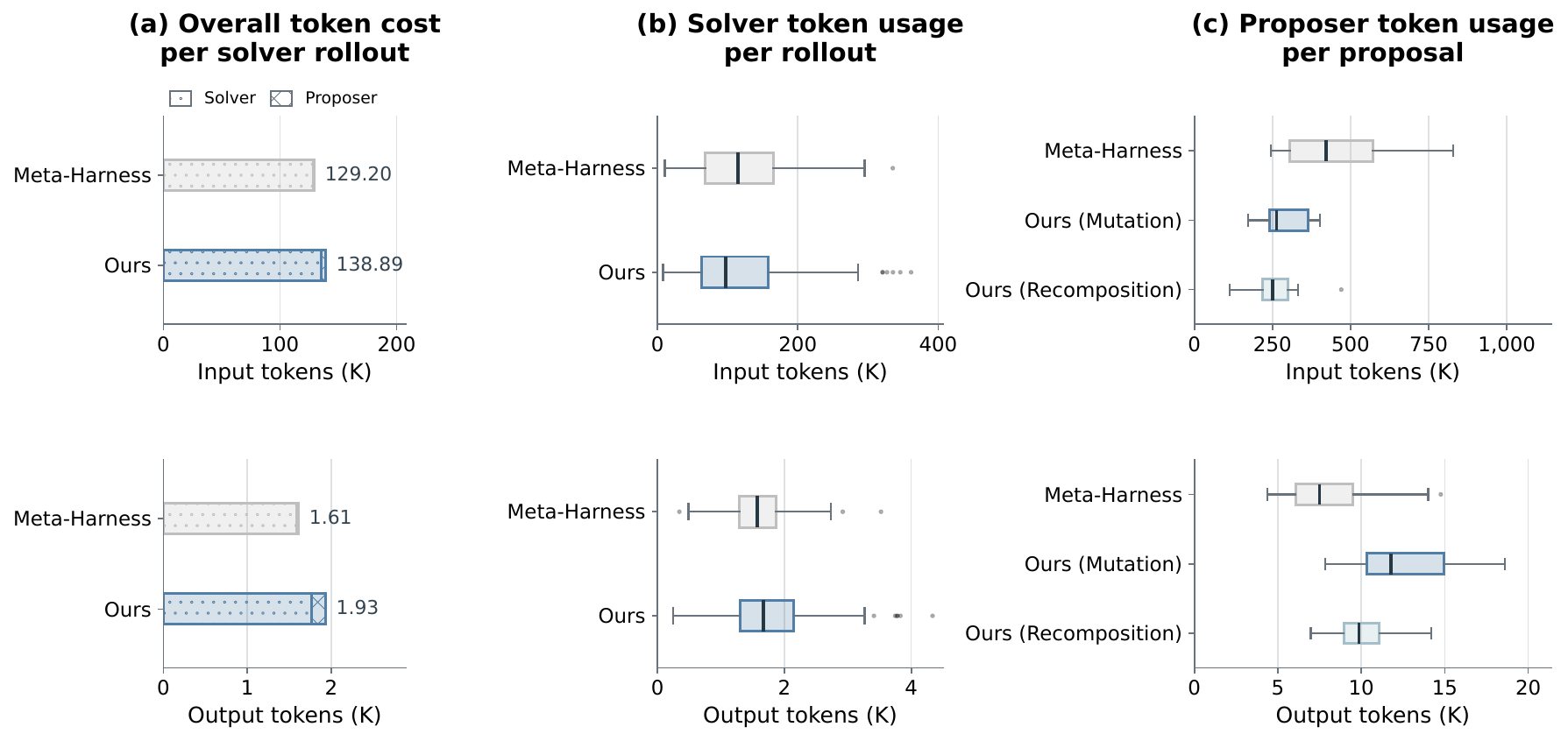}
    \caption{
        \textbf{Token cost analysis.}
        (a) Total token usage per solver rollout, including solver and proposer tokens.
        (b) Solver tokens per rollout.
        (c) Proposer tokens per proposal, with plugin mutation and harness recomposition reported separately.
    }
    \label{fig:analysis-cost}
\end{figure}

Figure~\ref{fig:analysis-cost} compares token usage between \plugharness and Meta-Harness.
We report total tokens per solver rollout, including proposer usage, and separately examine solver and proposer costs.
For \plugharness, proposer costs are further divided into plugin mutation and harness recomposition.
Total input and output tokens per solver rollout increase by 7.5\% and 19.8\%, respectively.
Solver token usage remains similar, while the additional mutation stage contributes to the proposer overhead.

\end{document}